\documentclass[11pt]{article}

\usepackage[margin=1.0in]{geometry}
\usepackage[T1]{fontenc}
\usepackage[utf8]{inputenc}
\usepackage{times}
\usepackage{amsmath,amssymb,amsthm,mathtools}
\usepackage{booktabs}
\usepackage{multirow}
\usepackage{adjustbox}
\usepackage{array}
\usepackage{graphicx}
\usepackage{subcaption}
\usepackage{float}
\usepackage{algorithm}
\usepackage{algpseudocode}
\usepackage[numbers,compress]{natbib}
\usepackage[colorlinks=true,linkcolor=blue!60!black,citecolor=blue!60!black,urlcolor=blue!60!black]{hyperref}
\usepackage{url}
\usepackage{parskip}
\usepackage{enumitem}
\usepackage{xcolor}
\usepackage{titlesec}
\usepackage{setspace}

\titleformat{\section}{\large\bfseries}{\thesection}{1em}{}
\titleformat{\subsection}{\normalsize\bfseries}{\thesubsection}{1em}{}
\titleformat{\paragraph}[runin]{\normalsize\bfseries}{}{0em}{}[.\quad]

\title{\textbf{Relative Repairability: A Calibration-Based Diagnostic for
High-Sparsity Post-Pruning Allocation}}

\author{
Qishi Zhan$^{1,\ast}$ \quad
Liang He$^{2,\ast}$ \quad
Minxuan Hu$^{3}$ \quad
Ziheng Chen$^{4}$ \\[4pt]
$^{1}$Marquette University \quad
$^{2}$Tongji University \quad
$^{3}$Cornell University \quad
$^{4}$UT Austin \\[2pt]
\texttt{qishi.zhan@marquette.edu, heliang19823@gmail.com} \\[2pt]
$^\ast$Corresponding authors
}
\date{}

\begin{document}

\maketitle

\begin{abstract}
At very high sparsity, neural network pruning does more than decide which
weights remain. It also determines where pruning induced damage is placed
across the network, and whether that damage can be recovered by a fixed
lightweight repair procedure. We study this problem through the lens of
repair conditioned sparsity allocation. We introduce Relative Repairability
(RR), a calibration based diagnostic that compares the raw activation
distortion caused by layerwise pruning with the residual distortion left after
channelwise variance matching repair. RR estimates the fraction of local
damage that remains after repair, using only unlabeled calibration data.
Across ResNet18, ResNet34, and VGG16 BN on CIFAR10 and CIFAR100, we find that
RR is not a universally dominant allocation rule. Instead, it is most useful
near an architecture dependent recoverability transition, where standard
structural or magnitude based allocation priors begin to lose reliability but
post repair recovery has not yet fully collapsed. On CIFAR100 ResNet18, a
fine grained sweep shows that RR improves over ERK across the central
transition band and surpasses LAMP near the upper part of this band. A
projection forced ablation further shows that capped ERK can over protect
projection layers, shifting excessive sparsity onto regular convolutions and
reducing post repair recovery. These results suggest that high sparsity
pruning should allocate not only retained weights, but also repairable damage.
\end{abstract}

\section{Introduction}
\label{sec:intro}

Post-training pruning is one of the most direct ways to compress a
trained neural network. A sparse model can be obtained without changing
the architecture and without retraining from scratch, making pruning
attractive when a dense model must be deployed under memory or compute
constraints~\cite{han2015learning,blalock2020state}. Yet at very high
sparsity, especially above 90\%, pruning failure is not fully described
by how many weights are removed. The same global sparsity can induce
very different forms of representational damage, and these differences
determine whether a subsequent lightweight repair procedure can recover
the sparse model.

This distinction matters because two masks with the same global sparsity
can have sharply different post-repair recoverability. If pruning assigns
too much damage to layers whose induced activation distortion cannot be
corrected by lightweight calibration, post-pruning repair may fail even
when the same number of parameters is retained. Conversely, a mask that
places more sparsity in layers whose distortion is largely repairable may
remain recoverable at the same global sparsity. High-sparsity pruning is
therefore not only a parameter selection problem, but also a problem of
allocating repairable damage across layers.

Existing sparsity allocation rules capture different parts of this
problem. Global magnitude pruning~\cite{han2015learning} is simple and
often effective at moderate sparsity, but it does not explicitly control
how sparsity is distributed across layers. Uniform layerwise pruning
imposes equal sparsity on each layer, but ignores layer sensitivity and
network structure. ERK allocation~\cite{evci2020rigl} provides a
structural prior by assigning density according to layer shape, while
LAMP~\cite{lee2021layeradaptive} provides a strong layer-adaptive
magnitude rule. However, neither structural density nor magnitude
normalization asks whether the damage assigned to a layer is actually
repairable by the chosen post-pruning correction method.
Activation-aware diagnostics~\cite{sun2024wanda,frantar2023sparsegpt}
address part of this issue by measuring pruning-induced activation
shift, but raw shift alone does not distinguish distortion that can be
corrected from distortion that remains after repair.

We propose \emph{Relative Repairability} (RR), a calibration-based
diagnostic for sparsity allocation under post-pruning repair. For each
layer and candidate sparsity, we measure the raw activation shift induced
by pruning and the residual activation shift after a lightweight
channelwise variance-matching correction. Their ratio estimates the
fraction of pruning-induced distortion that remains after repair. A low
ratio indicates that the layer's pruning damage is largely repairable
under the fixed repair operator, while a high ratio indicates that the
damage persists after correction. We use this diagnostic as a layerwise
allocation score, assigning more sparsity to layers whose pruning-induced
damage is more repairable and protecting layers whose residual distortion
remains high.

Our experiments show that RR is not a universal replacement for
structural or magnitude-based allocation. Instead, it is most informative
near a recoverability transition: a high-sparsity regime where standard
allocation rules begin to lose accuracy, but the sparse network has not
yet fully collapsed. We use ResNet18 as the main setting to characterize
this transition in detail. On CIFAR100, a fine-grained sweep from
93.5\% to 96.0\% sparsity shows that RR improves over ERK throughout a
contiguous 94.0--95.5\% transition band and becomes stronger than LAMP
near the upper part of this band. This demonstrates that the effect is not
a single selected sparsity point and not merely an artifact of comparing
against ERK.

We further analyze why RR changes the allocation in this regime. A
projection-forced ERK ablation shows that capped ERK can over-protect
projection layers, shifting excessive sparsity onto regular
convolutions and reducing post-repair recovery. RR identifies a different
damage placement from calibration response: it assigns more sparsity to
projection layers whose distortion is repairable and protects selected
regular convolutions whose residual distortion remains high.

Additional VGG16-BN and ResNet34 experiments support an
architecture-dependent boundary interpretation. On VGG16-BN, RR becomes
most informative near the collapse boundary, whereas ResNet34 remains
stable under ERK and LAMP until more extreme sparsity. Thus, the useful
region of RR should not be tied to a fixed numerical sparsity level.
Rather, RR acts as a repair-aware diagnostic for architectures that are
approaching collapse but still admit post-repair recovery.

Our contributions are as follows:
\begin{itemize}
    \item We formulate high-sparsity post-pruning allocation as a
    repairability-aware damage placement problem. Rather than asking only
    which layers should retain more parameters, we ask which
    pruning-induced damage can be recovered by a fixed lightweight repair
    operator.

    \item We introduce Relative Repairability (RR), a calibration-based
    diagnostic that normalizes post-repair activation residual by raw
    pruning-induced activation shift. RR estimates the fraction of local
    damage that remains after repair and uses this signal to allocate
    sparsity under a fixed global budget.

    \item We show that RR is most useful near an architecture-dependent
    recoverability boundary. On CIFAR100 ResNet18, RR improves over ERK
    across a contiguous transition band and exceeds LAMP in the upper part
    of that band. VGG16-BN and ResNet34 further show that the location of
    this boundary shifts with architecture.

    \item We provide a mechanism analysis of capped ERK on ResNet18,
    showing that over-protection of projection layers can shift excessive
    sparsity onto regular convolutions and reduce post-repair recovery.
\end{itemize}

\section{Related Work}
\label{sec:related}

\paragraph{Pruning and sparsity allocation.}
A large body of work removes parameters from trained neural networks
using magnitude or saliency criteria~\cite{han2015learning,
lecun1989optimal,hassibi1993optimal}. More recent post-training methods
use richer local information to reduce pruning error, including
second-order updates~\cite{frantar2023sparsegpt} and activation-aware
weight ranking~\cite{sun2024wanda}. Layerwise sparsity allocation has
also been studied directly. LAMP~\cite{lee2021layeradaptive} derives
layer-adaptive magnitude scores by controlling model-level
\(\ell_2\) distortion, while ERK~\cite{evci2020rigl} assigns density
according to a shape-based structural prior. We use LAMP and ERK as
complementary baselines: LAMP tests whether RR adds information beyond
layer-adaptive magnitude normalization, while ERK provides a structural
allocation rule whose projection-layer behavior admits a mechanistic
analysis.

\paragraph{Calibration-based post-pruning repair.}
Our work focuses on the post-training high-sparsity setting, where the
dense model is fixed, no retraining is performed, and only a small
unlabeled calibration set is available. A separate line of work studies
repair or recalibration under similar constraints. BatchNorm
recalibration~\cite{ioffe2015batch,li2017pruning} updates running
statistics after pruning and can recover a substantial portion of lost
accuracy. Recent work on signal collapse shows that one-shot pruning can
induce progressive activation-variance decay across layers, and
REFLOW~\cite{saikumar2025reflow} addresses this by restoring layerwise
activation variance from calibration data without gradient updates.
Related calibration-based corrections also appear in model merging, such
as REPAIR~\cite{jordan2023repair}, and in post-training
quantization~\cite{nagel2020adaround,hubara2021accurate} and
pruning~\cite{hubara2021accelerated,lazarevich2021post}. We adopt the
same no-retraining philosophy, but use repair not only as a post-hoc
correction step: we treat the repair response itself as a diagnostic for
sparsity allocation.

\paragraph{From activation shift to repairability.}
Activation-aware pruning methods~\cite{sun2024wanda,frantar2023sparsegpt}
use calibration inputs to estimate the effect of pruning on intermediate
features. These methods measure the size of the initial perturbation, but
do not ask whether that perturbation can be corrected by a specified
post-pruning repair operator. RR instead compares the raw activation
shift with the residual shift left after channelwise repair. This makes
the diagnostic conditional on the chosen repair procedure: allocation is
guided not only by how much damage pruning causes, but by how much of
that damage remains after repair.

\section{Method: Repairability-Guided Sparsity Allocation}
\label{sec:method}

We describe Relative Repairability (RR), a calibration-based diagnostic
for allocating sparsity under a fixed post-pruning repair operator. The
goal is not to predict final accuracy exactly, but to identify where
pruning-induced damage is more compatible with lightweight repair.

\subsection{Post-pruning repair setting}
\label{sec:setting}

Let \(f_\theta\) be a dense convolutional network with prunable layers
\(\ell=1,\ldots,L\). The dense weights are fixed, a target global
sparsity \(S\) is specified, and only a small unlabeled calibration set
\(\mathcal C\) is available. For layer \(\ell\), let \(n_\ell\) denote
the number of prunable parameters and \(s_\ell\in[0,1]\) its assigned
sparsity. The achieved global sparsity is
\begin{equation}
S(\mathbf{s})
=
\frac{\sum_{\ell=1}^{L} n_\ell s_\ell}
     {\sum_{\ell=1}^{L} n_\ell}.
\label{eq:global_sparsity}
\end{equation}

To estimate local pruning damage, we use diagnostic models in which one
layer \(\ell\) is pruned to candidate sparsity \(s\), while all other
layers remain dense. These models are used only to measure calibration
activations. The final sparse network is obtained by pruning all layers
according to the selected allocation and then applying the same
lightweight repair pipeline.

Throughout the paper, the fixed repair operator is channelwise
variance-matching correction followed by BatchNorm recalibration,
abbreviated as CR+BN. The repair operator uses calibration activations
to rescale output channels toward their dense activation variance, with
shrinkage for collapsed channels to avoid noise amplification. Full
details are provided in Appendix~\nameref{app:repair_details}. RR should
therefore be interpreted as a repair-operator-conditioned diagnostic,
not as an intrinsic repair-independent layer property.

\subsection{Relative repairability}
\label{sec:relative_repairability}

For layer \(\ell\), let
\(a_\ell(x)\in\mathbb{R}^{C\times H\times W}\) denote the dense
post-convolution activation on calibration input \(x\), and let
\(a_{\ell,s}^{\mathrm{pruned}}(x)\) denote the activation after pruning
only layer \(\ell\) to sparsity \(s\). We define the raw
pruning-induced activation distortion as
\begin{equation}
D_{\ell}^{\mathrm{raw}}(s)
=
\frac{1}{|\mathcal C|}
\sum_{x\in\mathcal C}
d\!\left(a_\ell(x),a_{\ell,s}^{\mathrm{pruned}}(x)\right),
\label{eq:raw_shift}
\end{equation}
where
\begin{equation}
d(a,b)
=
\frac{1}{C}
\sum_{c=1}^{C}
\frac{\|a_c-b_c\|_2^2}{\|a_c\|_2^2+\epsilon}.
\label{eq:distance}
\end{equation}
Let \(a_{\ell,s}^{\mathrm{repair}}(x)\) denote the activation after
applying CR+BN to the single-layer pruned diagnostic model. The
post-repair residual distortion is
\begin{equation}
D_{\ell}^{\mathrm{repair}}(s)
=
\frac{1}{|\mathcal C|}
\sum_{x\in\mathcal C}
d\!\left(a_\ell(x),a_{\ell,s}^{\mathrm{repair}}(x)\right).
\label{eq:repair_residual}
\end{equation}
Relative Repairability is the normalized residual ratio
\begin{equation}
R_{\ell}(s)
=
\frac{D_{\ell}^{\mathrm{repair}}(s)+\epsilon}
     {D_{\ell}^{\mathrm{raw}}(s)+\epsilon}.
\label{eq:repair_ratio}
\end{equation}

Raw shift measures the size of the initial pruning perturbation, while
the unnormalized repair residual mixes the perturbation size with the
fraction left uncorrected. RR separates these effects by estimating the
fraction of pruning-induced distortion that remains after repair. Lower
values indicate more repairable damage; higher values indicate
repair-resistant damage. Values above one are possible and indicate that
the repair operator increases the measured discrepancy.

\subsection{Diagnostic-based allocation}
\label{sec:allocation}

For each layer, we evaluate a candidate sparsity grid
\(\mathcal S=\{s_1<\cdots<s_K\}\). Given a diagnostic score
\(g(\ell,s)\), such as raw shift, repair residual, or RR, we view
allocation as the discrete budgeted problem
\begin{equation}
\min_{\{s_\ell\in\mathcal S\}_{\ell=1}^{L}}
\sum_{\ell=1}^{L} g(\ell,s_\ell)
\quad
\text{subject to}
\quad
\frac{\sum_{\ell=1}^{L} n_\ell s_\ell}
     {\sum_{\ell=1}^{L} n_\ell}
\ge S.
\label{eq:allocation_objective}
\end{equation}
This objective is a diagnostic proxy rather than an exact model of final
accuracy, because scores are measured from single-layer perturbations
while the final model prunes all layers simultaneously. 
We validate this local diagnostic proxy in
Appendix~\nameref{app:local_joint_consistency}, where the allocation-level
RR objective is shown to be positively rank-correlated with final
CR+BN recovery across allocation rules in the CIFAR100 ResNet18
transition band.

We solve this problem with a common greedy solver for all diagnostics.
Starting from the lowest candidate sparsity, the solver repeatedly
promotes the layer whose next sparsity step has the smallest diagnostic
cost increase per additionally pruned parameter:
\begin{equation}
q_\ell
=
\frac{
g(\ell,s_{k_\ell+1})-g(\ell,s_{k_\ell})
}{
(s_{k_\ell+1}-s_{k_\ell})n_\ell
}.
\label{eq:greedy_increment}
\end{equation}
Using the same solver for raw shift, repair residual, and RR ensures
that differences in recovered accuracy are attributable to the
diagnostic signal rather than to the allocation procedure. Full
pseudocode and implementation details are given in
Appendix~\nameref{app:allocation_details}.

\section{Experimental Setup}
\label{sec:setup}

We evaluate RR in a post-training, label-free repair setting. The dense
model is fixed before pruning, and no gradient updates are used during
diagnosis or repair. Unless otherwise stated, all reported accuracies are
top-1 test accuracies after the full CR+BN repair pipeline.

\paragraph{Datasets and architectures.}
Our main experiments use ResNet18~\cite{he2016deep} on CIFAR10 and
CIFAR100. ResNet18 is initialized from ImageNet-pretrained weights and
fine-tuned on the target dataset for five epochs with frozen lower
layers, following a controlled transfer-learning setup. We use this
setting for the main sparsity sweep, the fine-grained transition
analysis, and the ERK mechanism ablation. To test whether the
recoverability boundary shifts with architecture, we additionally
evaluate VGG16-BN and ResNet34 on CIFAR10 and CIFAR100; full
architecture-specific results are reported in
Appendix~\nameref{app:vgg_results} and Appendix~\nameref{app:resnet34_results}.

\paragraph{Calibration and repair.}
All activation diagnostics and channelwise repair statistics are
estimated from 128 unlabeled training images, collected as two batches of
64. The same calibration examples are reused across diagnostics within a
seed to keep comparisons controlled. After channelwise repair, BatchNorm
running statistics are recalibrated using 20 batches of 128 unlabeled
training images with the default running-statistics momentum. No labels,
gradients, or retraining are used during repair.

\paragraph{Allocation methods.}
We compare RR with global magnitude pruning, uniform layerwise pruning,
raw activation shift allocation, unnormalized repair residual
allocation, ERK~\cite{evci2020rigl}, and LAMP~\cite{lee2021layeradaptive}.
Global magnitude and uniform pruning provide simple allocation
baselines. ERK provides a structural density prior, while LAMP provides a
strong layer-adaptive magnitude baseline. Raw shift and repair residual
isolate the contribution of activation diagnostics from the ratio
normalization in RR.

\paragraph{Candidate sparsities.}
For diagnostic-based allocation methods, we use the candidate sparsity
grid
\[
\mathcal{S}
=
\{0.70,0.80,0.85,0.90,0.925,0.95,0.975\}.
\]
The greedy allocation procedure stops once the target global sparsity is
reached. Following the convention used in post-pruning repair, the first
convolutional layer is kept dense and excluded from diagnostic
allocation.

\paragraph{Sparsity sweeps and seeds.}
The main ResNet18 sweep evaluates target sparsities of 90\%, 92.5\%,
95\%, and 97.5\% on both CIFAR10 and CIFAR100, averaged over three
random seeds. We additionally run a fine-grained CIFAR100 ResNet18 sweep
from 93.5\% to 96.0\% in 0.5 percentage-point increments to test whether
RR's advantage is a single selected sparsity point or a contiguous
transition regime. LAMP is included on the central 94.0--95.5\%
transition band, where the comparison with a strong magnitude-based
allocation baseline is most informative. Numerical stability constants
are set to \(\epsilon=10^{-8}\) throughout.

\paragraph{ERK implementation.}
For ERK, we use the standard density rule with maximum density cap
\(1.0\) and minimum density floor \(0.025\). At extreme sparsity, some
low-parameter projection layers can receive uncapped ERK density above
one and are therefore retained at full density after clipping. We report
this detail because it directly motivates the projection-layer mechanism
analysis in Sec.~\nameref{sec:refines_erk}.

\section{Experimental Results}
\label{sec:results}

\subsection{Layerwise allocation controls post-repair recoverability}
\label{sec:main_results}

\begin{table*}[t]
\centering
\small
\begin{tabular}{lccccccc}
\toprule
Dataset & Sparsity & Global & Uniform & Raw Shift & Repair Residual & RR & ERK \\
\midrule
CIFAR100 & 90\%   & 15.39 & 23.97 & 35.93 & 35.82 & 37.59          & \textbf{38.47} \\
CIFAR100 & 92.5\% & 8.77  & 9.85  & 19.84 & 19.90 & 23.90          & \textbf{24.12} \\
CIFAR100 & 95\%   & 3.80  & 3.54  & 10.00 & 9.87  & \textbf{11.86} & 9.38 \\
CIFAR100 & 97.5\% & 1.64  & 1.61  & 1.64  & 1.63  & 1.64           & 1.64 \\
\midrule
CIFAR10  & 90\%   & 63.26 & 55.39 & 77.09 & \textbf{79.12} & 77.37 & 77.14 \\
CIFAR10  & 92.5\% & 42.92 & 36.37 & 60.22 & 60.61 & \textbf{66.26} & 63.88 \\
CIFAR10  & 95\%   & 25.54 & 22.97 & 38.91 & 38.89 & \textbf{44.07} & 37.89 \\
CIFAR10  & 97.5\% & 12.94 & 11.88 & 11.98 & 11.88 & 11.95          & 11.95 \\
\bottomrule
\end{tabular}
\caption{Recovered accuracy (\%) after CR+BN on ResNet18, averaged over
three seeds. Best per row is bolded except at 97.5\%, where all methods
are within seed-level noise of chance-level accuracy and the small
differences should not be over-interpreted.}
\label{tab:main_results}
\end{table*}

\begin{figure*}[t]
\centering
\includegraphics[width=\linewidth]{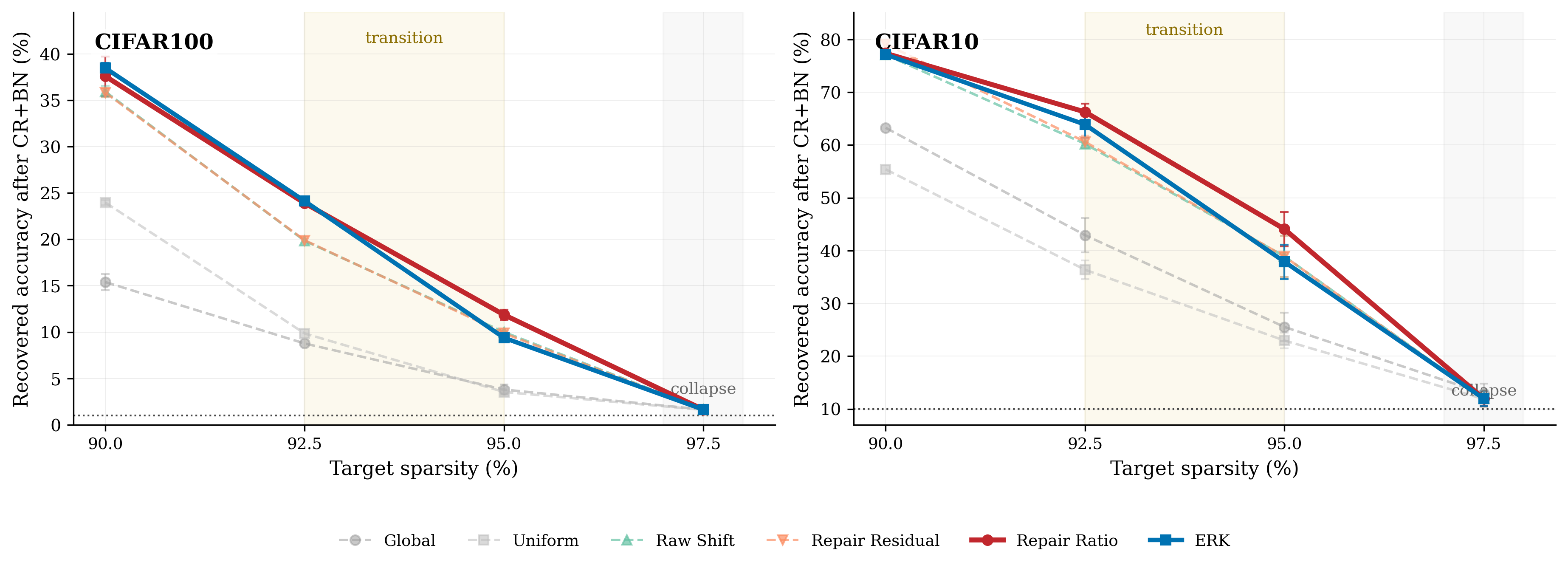}
\caption{Sparsity sweep on CIFAR100 and CIFAR10 with ResNet18. All
points are averaged over three seeds, with error bars indicating
standard deviation. RR is competitive at moderate high sparsity, becomes
strongest in the transition regime, and converges with all baselines once
the model crosses the recoverability threshold.}
\label{fig:sparsity_sweep}
\end{figure*}

Table~\ref{tab:main_results} and Fig.~\ref{fig:sparsity_sweep} show that layerwise allocation
strongly controls post-repair recoverability. At the same target global
sparsity, different allocation rules produce substantially different
recovered accuracies after CR+BN. On both datasets, global and uniform
pruning degrade rapidly as sparsity increases, while nonuniform
allocation rules remain much more recoverable in the 90\%--95\% range.
Thus, high-sparsity failure is not determined by the number of retained
weights alone; where pruning damage is placed across layers has a large
effect on whether lightweight repair can recover the sparse model.

The same table also shows that RR is not uniformly dominant. On
CIFAR100, ERK is slightly stronger at 90\% and 92.5\%, while RR becomes
strongest at 95\%. On CIFAR10, the transition appears earlier: RR is
competitive at 90\%, becomes strongest at 92.5\%, and remains strongest
at 95\%. At 97.5\%, all methods approach chance-level accuracy, showing
that once the model has crossed the recoverability boundary, allocation
signals have little room to help.

\subsection{RR is most useful in a contiguous transition band}
\label{sec:transition}

\begin{figure}[t]
\centering
\includegraphics[width=0.7\linewidth]{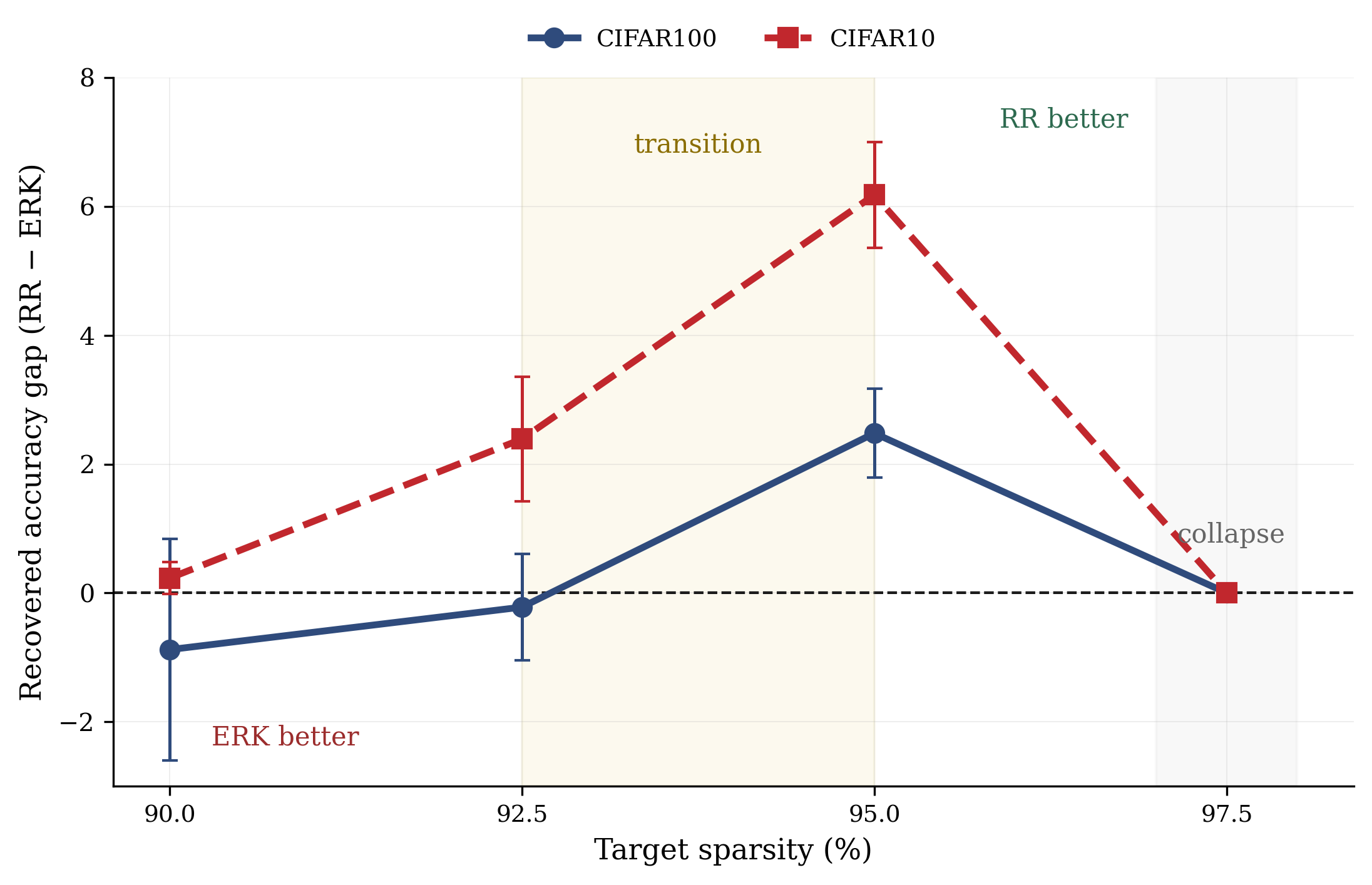}
\caption{Recovered accuracy gap between RR and ERK across sparsity
levels. Error bars show the standard deviation of the paired RR minus ERK
accuracy gap across three seeds. The advantage of RR emerges in the
recoverability transition regime and vanishes once all methods collapse
near chance-level performance.}
\label{fig:rr_erk_gap}
\end{figure}

The main sparsity sweep suggests that the benefit of RR is concentrated
near the recoverability transition. At moderate high sparsity, structural
and activation-based allocation rules are already effective, and RR is
competitive rather than uniformly dominant. In the transition regime, RR
becomes more useful: at 95\%, it improves over ERK by 2.48 points on
CIFAR100 and 6.18 points on CIFAR10. Once accuracy approaches chance
level at 97.5\%, the RR--ERK gap vanishes within seed-level noise
(Fig.~\ref{fig:rr_erk_gap}).

To test whether this effect is a single selected sparsity point, we run a
fine-grained CIFAR100 ResNet18 sweep from 93.5\% to 96.0\% in 0.5
percentage-point increments. We also include LAMP on the central
94.0--95.5\% band, where comparison with a strong magnitude-based
allocation baseline is most informative.

\begin{table}[t]
\centering
\footnotesize
\setlength{\tabcolsep}{2.8pt}
\begin{tabular}{ccccc}
\toprule
Sp. & RR & LAMP & ERK & RR$-$LAMP \\
\midrule
94.0\% & \(17.68{\pm}0.35\) & \(17.43{\pm}0.46\) & \(14.56{\pm}0.79\) & \(+0.25\) \\
94.5\% & \(14.20{\pm}0.55\) & \(14.19{\pm}0.57\) & \(11.61{\pm}0.32\) & \(+0.01\) \\
95.0\% & \(11.88{\pm}0.49\) & \(10.76{\pm}0.15\) & \(9.38{\pm}0.15\)  & \(+1.12\) \\
95.5\% & \(9.55{\pm}0.63\)  & \(7.93{\pm}0.30\)  & \(7.23{\pm}0.40\)  & \(+1.62\) \\
\bottomrule
\end{tabular}
\caption{Central CIFAR100 ResNet18 transition band. Results are
recovered accuracy after CR+BN, averaged over three seeds. LAMP is a
strong layer-adaptive magnitude baseline; RR remains competitive at
94.0--94.5\% and becomes stronger near the upper transition region.}
\label{tab:fine_transition_lamp}
\end{table}

\begin{figure}[t]
\centering
\includegraphics[width=0.7\linewidth]{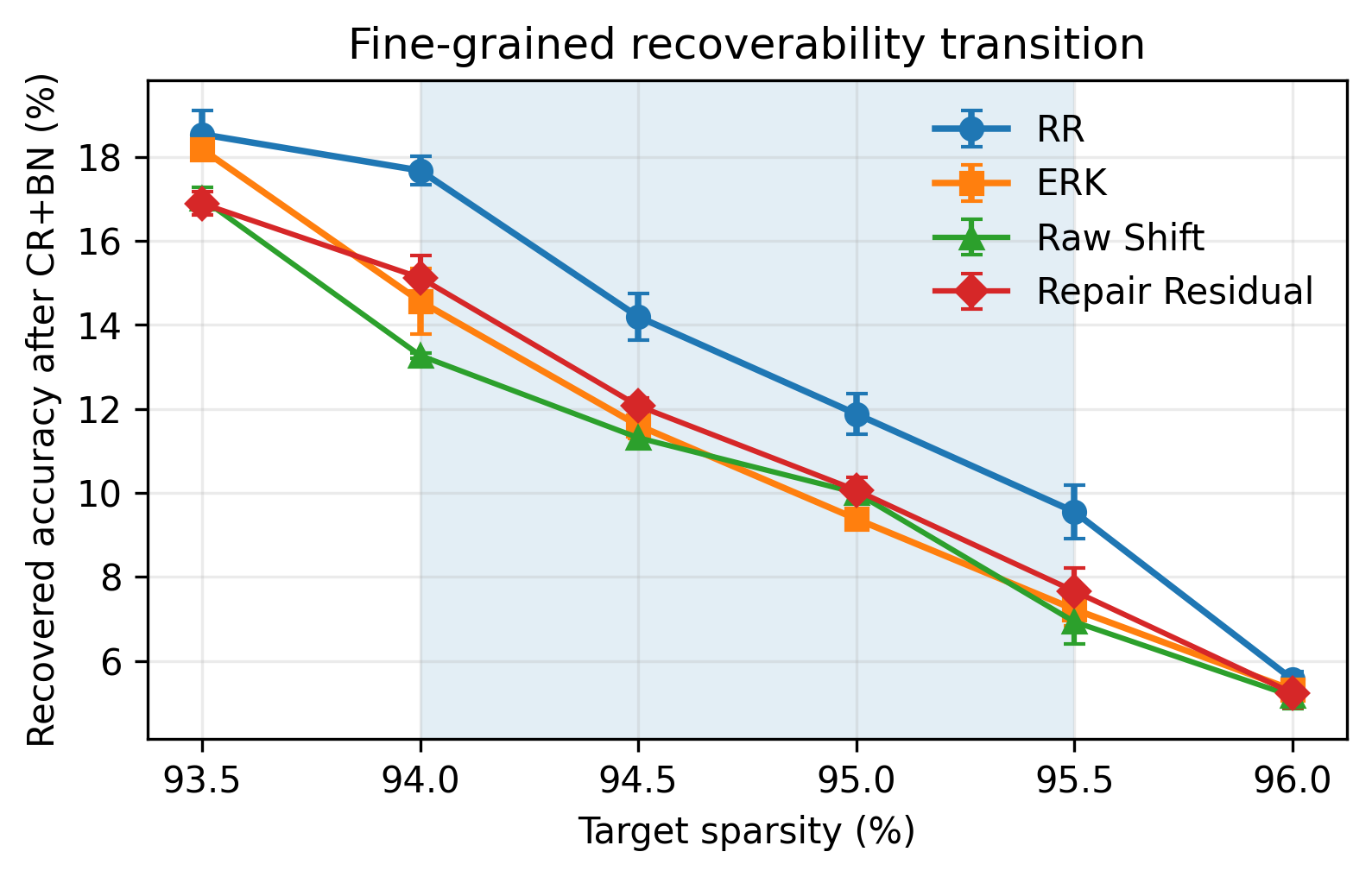}
\caption{Fine-grained recoverability transition on CIFAR100 ResNet18
with LAMP included. The shaded region marks the central transition band.
LAMP is stronger than ERK, but RR remains competitive with LAMP and
improves over it as sparsity approaches the upper transition region.}
\label{fig:fine_transition}
\end{figure}

Table~\ref{tab:fine_transition_lamp} and Fig.~\ref{fig:fine_transition} show that the
advantage of RR is not confined to a single 95\% sparsity point. Against
ERK, RR improves by \(3.12 \pm 0.64\), \(2.59 \pm 0.78\),
\(2.50 \pm 0.63\), and \(2.32 \pm 0.24\) points at 94.0\%, 94.5\%,
95.0\%, and 95.5\%, respectively. The gain is much smaller at 93.5\%
and 96.0\%, supporting a transition-regime interpretation: below the
transition, classical allocation rules remain competitive; within the
transition, repair-aware damage placement substantially affects recovery;
near collapse, all allocation rules have limited room to help.

LAMP strengthens this interpretation. It is substantially stronger than
ERK throughout the central transition band, showing that ERK should not
be viewed as the strongest competing allocation rule. Nevertheless, RR
remains competitive with LAMP at 94.0--94.5\% and improves over it by
1.12 and 1.62 points at 95.0\% and 95.5\%, respectively. RR also
improves over raw activation shift and unnormalized repair residual
throughout the central transition band, indicating that the ratio
normalization provides information beyond either the initial activation
perturbation or the absolute post-repair residual.

\subsection{Calibration-only identification of the transition core}
\label{sec:calibration_transition_detector}

The fine-grained sparsity sweep in Sec.~\nameref{sec:transition} shows that the advantage of RR is concentrated in a recoverability transition band. 
However, if this band is identified only from recovered test accuracy, one may ask whether the transition region is selected post hoc. 
We therefore introduce a calibration-only transition detector that uses only RR diagnostic curves before evaluating the final sparse model.

For a target sparsity $s$, let $A^{\mathrm{RR}}_s$ and $A^{\mathrm{ERK}}_s$ denote the layerwise allocations produced by RR and ERK, respectively. 
Given the RR diagnostic curve $R_{\ell}(\cdot)$ for each layer, we define the RR objective value of an allocation as
\[
J_{\mathrm{RR}}(A_s)
=
\frac{1}{L}
\sum_{\ell=1}^{L}
R_{\ell}(s_{\ell}),
\]
where $s_{\ell}$ is the sparsity assigned to layer $\ell$ by allocation $A_s$. 
This objective does not use labels or test accuracy; it only measures how much repair-resistant distortion remains according to the calibration diagnostic.

We then compute the RR-objective gap between ERK and RR:
\[
G(s)
=
J_{\mathrm{RR}}(A^{\mathrm{ERK}}_s)
-
J_{\mathrm{RR}}(A^{\mathrm{RR}}_s).
\]
A large $G(s)$ indicates that, under the RR diagnostic, ERK places more repair-resistant damage than the RR allocation. 
This gap alone may also be nonzero at lower sparsity, where the network is not yet under substantial repair stress. 
We therefore weight it by a normalized repair-stress term:
\[
T(s)
=
\mathrm{clip}
\left(
\frac{
J_{\mathrm{RR}}(A^{\mathrm{RR}}_s)
-
J_{\mathrm{RR}}(A^{\mathrm{RR}}_{s_{\mathrm{low}}})
}{
J_{\mathrm{RR}}(A^{\mathrm{RR}}_{s_{\mathrm{high}}})
-
J_{\mathrm{RR}}(A^{\mathrm{RR}}_{s_{\mathrm{low}}})
+\epsilon
},
0,1
\right),
\]
where $s_{\mathrm{low}}=90\%$ and $s_{\mathrm{high}}=97.5\%$ are anchor sparsities. 
The calibration transition score is
\[
\mathrm{CTS}(s)=G(s)\cdot T(s).
\]
Intuitively, CTS becomes large only when two conditions hold simultaneously: ERK and RR disagree under a repairability objective, and the model is already under nontrivial repair stress.

We smooth CTS using a three-point moving average, denoted by $\overline{\mathrm{CTS}}(s)$. 
The high-confidence transition core is defined as the longest contiguous sparsity interval satisfying
\[
\overline{\mathrm{CTS}}(s)
\ge
B+\alpha
\left(
\max_{s'}\overline{\mathrm{CTS}}(s')-B
\right),
\]
where $B$ is the maximum smoothed score over the low-sparsity anchors and we set $\alpha=0.7$ throughout. 
In addition to this high-confidence core, we also record a broader detected band to show the wider region where the calibration transition score is elevated. 
The detector is not intended to predict final accuracy exactly. 
Its purpose is to identify, before test evaluation, the sparsity region where repair-aware allocation is expected to matter.
In the reported settings, the detected high-confidence cores align well with the
sparsity regions where RR becomes strongest or most competitive in recovered accuracy.

Table~\ref{tab:cts_detection} summarizes the calibration-only detection results. 
On CIFAR100 ResNet18, the detector selects $95.0$--$95.5\%$ as the high-confidence transition core. 
This interval is the upper core of the accuracy-based transition band in Sec.~\nameref{sec:transition}, where RR not only improves over ERK but also surpasses LAMP. 
Thus, the most important part of the transition band is reflected in calibration diagnostics before recovered test accuracy is evaluated.

The detector also captures architecture-dependent boundary shifts. 
For VGG16-BN, the selected high-confidence core appears near its collapse boundary. 
On CIFAR10, two seeds select $93.5$--$95.5\%$, while one seed selects $92.0$--$93.5\%$; on CIFAR100, all seeds select $94.0$--$95.5\%$. 
This is consistent with VGG16-BN approaching its recoverability boundary in the mid-to-high 90\% sparsity range. 
For ResNet34, the detected core moves further toward extreme sparsity. 
The detector selects $97.5\%$ on CIFAR10 and $97.0$--$97.5\%$ on CIFAR100, matching the later transition in which RR becomes most useful only near the collapse boundary.
For CIFAR10 ResNet18, the detector selects $97.0\%$, indicating that the calibration score identifies a late RR-useful high-stress region on this easier dataset.
Overall, these results support the view that RR is a transition-window diagnostic rather than a fixed-sparsity rule.

Full per-seed detection outputs are provided in Appendix~\nameref{app:cts_detection}. 
There, Table~\ref{tab:cts_detection_per_seed_band} reports the broad detected bands and high-confidence cores.
\begin{table*}[t]
\centering
\small
\setlength{\tabcolsep}{3.5pt}
\caption{Calibration-only transition detection. 
The detector uses only RR diagnostic curves and does not use test accuracy. 
We report the seed-stable or majority high-confidence core and its relation to the observed accuracy behavior.}
\label{tab:cts_detection}
\begin{tabular}{llcl}
\toprule
Architecture & Dataset & CTS-selected core & Accuracy behavior \\
\midrule
ResNet18 & CIFAR100 
& 95.0--95.5\% 
& Upper core where RR improves over ERK and surpasses LAMP \\

ResNet18 & CIFAR10
& 97.0\%
& Late high-stress region where RR remains useful before collapse \\

VGG16-BN & CIFAR10 
& 93.5--95.5\% 
& Near-collapse transition; one seed detects an earlier 92.0--93.5\% core\\

VGG16-BN & CIFAR100 
& 94.0--95.5\% 
& Transition near collapse boundary; RR strongest near 95\% \\

ResNet34 & CIFAR10 
& 97.5\% 
& Late transition; RR strongest at extreme sparsity \\

ResNet34 & CIFAR100 
& 97.0--97.5\% 
& Late repairability boundary; broad signals around 96.0--98.0\% \\
\bottomrule
\end{tabular}
\end{table*}

\subsection{Mechanism: RR refines capped ERK}
\label{sec:refines_erk}

\begin{figure}[h]
\centering
\includegraphics[width=0.7\linewidth]{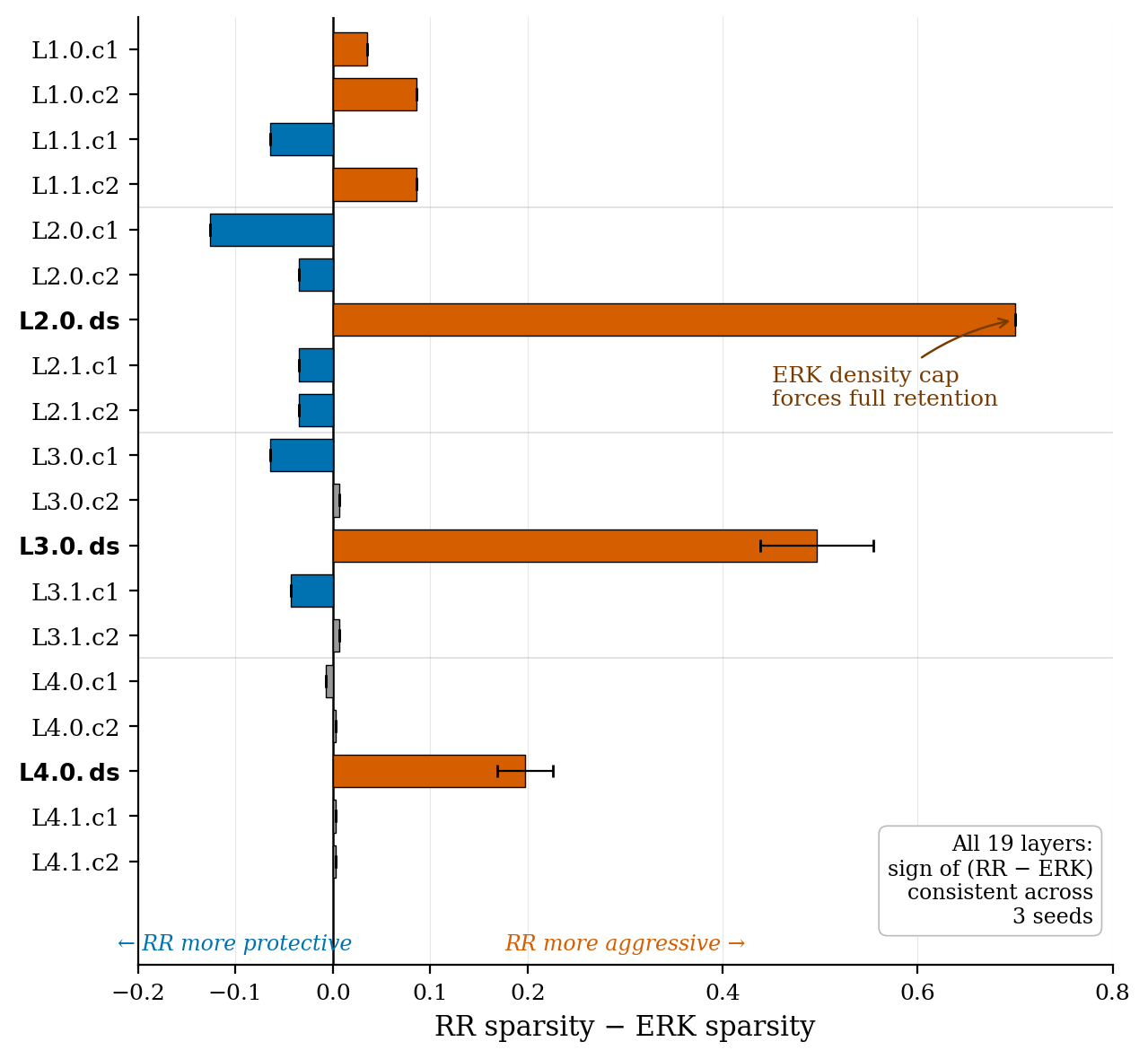}
\caption{Layerwise sparsity deviation between RR and ERK on CIFAR100
ResNet18 at 95\% sparsity. Positive values indicate layers where RR
assigns higher sparsity than ERK; negative values indicate layers it
protects more strongly. The largest deviations occur in projection
(downsample) layers, especially \texttt{layer2.0.downsample.0}, which
ERK retains entirely due to its density cap. Error bars show standard
deviation across three seeds.}
\label{fig:rr_erk_deviation}
\end{figure}

Although LAMP is stronger than ERK in the central CIFAR100 transition
band, ERK remains useful for mechanism analysis because its structural
density rule exposes a concrete failure mode. We therefore examine how
RR differs from ERK in layerwise allocation
(Fig.~\ref{fig:rr_erk_deviation}). We focus on CIFAR100 ResNet18 at 95\%
sparsity, where RR improves over both ERK and LAMP after CR+BN.

The two allocations are highly correlated, indicating that RR does not
discard the structural prior. However, their deviations are stable across
seeds. The largest deviations occur in projection layers. Under the
standard capped ERK rule, the uncapped density of
\texttt{layer2.0.downsample.0} exceeds one and is clipped, causing ERK
to retain this layer entirely. RR instead assigns this layer 70\%
sparsity while improving recovered accuracy. It also assigns higher
sparsity than ERK to other projection layers, while protecting selected
regular convolutions more strongly.

\begin{table}[t]
\centering
\small
\setlength{\tabcolsep}{3.8pt}
\begin{tabular}{cccc}
\toprule
Proj. sp. & Reg. conv sp. & BN & CR+BN \\
\midrule
0\%  & 96.49\% & \(3.51 \pm 0.89\) & \(5.63 \pm 0.39\) \\
70\% & 95.39\% & \(6.63 \pm 1.40\) & \(10.83 \pm 0.47\) \\
90\% & 95.08\% & \(6.99 \pm 1.62\) & \(11.55 \pm 0.59\) \\
\bottomrule
\end{tabular}
\caption{Projection-forced ERK ablation on CIFAR100 ResNet18 at 95\%
target sparsity. Projection layers are fixed to a prescribed sparsity,
and the remaining budget is reallocated to regular convolutions using
ERK. Dense projection layers force higher sparsity onto regular
convolutions and reduce post-repair recovery.}
\label{tab:projection_forced}
\end{table}

The projection-forced ablation in Table~\ref{tab:projection_forced} confirms
that this is not only a descriptive difference. Retaining projection
layers densely gives only \(5.63 \pm 0.39\%\) CR+BN recovery. Assigning
70\% projection sparsity raises recovery to \(10.83 \pm 0.47\%\), and
assigning 90\% projection sparsity gives \(11.55 \pm 0.59\%\). This does
not imply that projection sparsity should always be maximized. Rather,
it shows that the dense-projection solution induced by capped ERK is a
poor damage placement in this transition regime. By over-protecting
projection layers, ERK shifts excessive sparsity onto regular
convolutions, where the resulting damage is less recoverable under
CR+BN. RR discovers a related reallocation from calibration-based repair
response without manually specifying a projection-layer rule.

\subsection{Architecture-dependent boundary shifts}
\label{sec:architecture_boundary}

The ResNet18 experiments characterize the recoverability transition in
detail. We next ask whether this transition occurs at a fixed sparsity
level or shifts with architecture. We evaluate VGG16-BN and ResNet34 on
CIFAR10 and CIFAR100, and summarize the results in
Table~\ref{tab:arch_summary}. Full per-sparsity results, RR gap tables, and
achieved sparsities are reported in the appendix.

\begin{table*}[t]
\centering
\small
\setlength{\tabcolsep}{3.5pt}
\begin{tabular}{llcl}
\toprule
Architecture & Dataset & Region & Main pattern \\
\midrule
VGG16-BN & CIFAR10 & 92.5--95\% &
repair-aware rules become strongest \\
VGG16-BN & CIFAR100 & 95\% &
RR best near collapse \\
ResNet18 & CIFAR100 & 94.0--95.5\% &
RR improves over ERK; beats LAMP at upper band \\
ResNet34 & CIFAR10/100 & 97.5\% &
RR best only at extreme sparsity \\
\bottomrule
\end{tabular}
\caption{Architecture-dependent recoverability boundary. The table
summarizes where RR becomes most informative after CR+BN. Full
per-sparsity results are provided in the appendix.}
\label{tab:arch_summary}
\end{table*}

The additional architectures suggest that the recoverability boundary is
architecture-dependent. VGG16-BN, a plain feed-forward CNN, shows a
near-collapse transition: RR is weak at moderate sparsity but becomes
strongest near the 95\% collapse boundary. ResNet34 shows the
complementary pattern. ERK and LAMP remain strong through 95\%, while RR
becomes best only at the more extreme 97.5\% point. ResNet18 lies between
these cases, with a central transition band around 94.0--95.5\% on
CIFAR100.

These results are not intended to show that RR dominates across all
architectures and sparsities. Instead, they support the interpretation
that RR is most useful when a given architecture approaches its
recoverability boundary, where standard structural or magnitude-based
allocation priors begin to lose reliability but post-repair recovery has
not fully collapsed.

\section{Limitations}
\label{sec:limitations}

Our results support a regime-dependent view of repairability-guided
allocation, but the empirical scope remains limited. Although we evaluate
ResNet18, VGG16-BN, and ResNet34 on CIFAR-scale datasets, the study is
still confined to image classification under controlled transfer
settings. Larger datasets, additional model families, and
non-classification tasks are needed to determine how broadly
architecture-dependent recoverability boundaries generalize. The current
experiments also identify the recoverability transition through sparsity
sweeps.
Although Sec.~\nameref{sec:calibration_transition_detector} provides a calibration-only detector for the high-confidence transition core, CTS should be interpreted as an indicator of where repair-aware allocation is expected to matter rather than an exact predictor of final recovered accuracy. 
Extending such calibration-only indicators to predict the full transition band across broader architectures and tasks remains future work.

RR should also be interpreted with a restricted scope. 
It is not designed to replace structural or magnitude-based allocation rules uniformly across sparsity levels. 
Its advantage is concentrated near the recoverability transition, where standard allocation priors begin to lose reliability but the network has not fully collapsed. 
RR is conditioned on the chosen post-pruning repair operator. 
In this work, the operator is fixed to channelwise variance-matching repair followed by BatchNorm recalibration. 
Different repair operators may induce different repairability landscapes and therefore different allocations. 
Studying operator-agnostic repairability or jointly selecting the repair operator and allocation is left for future work. 
Thus, our claim is not that RR is a universal allocation rule, but that post-repair residuals provide a useful allocation signal under a fixed lightweight repair pipeline.

\section{Conclusion}
\label{sec:conclusion}

We studied high-sparsity post-pruning recovery as a
repairability-aware allocation problem. Rather than viewing pruning only
as the selection of retained weights, we asked where pruning-induced
damage can be placed so that a fixed lightweight repair operator can
still recover the sparse model. We introduced Relative Repairability
(RR), a calibration-based diagnostic that estimates the fraction of raw
activation distortion that remains after channelwise repair and uses
this signal for layerwise sparsity allocation.

Our experiments show that RR is most useful near the recoverability
transition, rather than uniformly across all sparsity levels. On
ResNet18, RR improves over ERK across a contiguous CIFAR100 transition
band and becomes stronger than LAMP near the upper part of the band. A
projection-forced ERK ablation further shows that capped ERK can
over-protect projection layers, shifting excessive sparsity onto regular
convolutions and reducing post-repair recovery. Additional VGG16-BN and ResNet34 experiments suggest that the location of
this boundary is architecture-dependent: VGG16-BN becomes informative
near its collapse boundary, while ResNet34 reaches the RR-useful region
only at more extreme sparsity.

These findings suggest that effective high-sparsity pruning should
allocate not only retained weights, but also repairable damage. RR
provides one way to expose this repairability structure under a fixed
calibration-based repair operator. Its role is therefore best understood
as a repair-aware diagnostic for the recoverability transition, rather
than as a universal sparsity allocation rule.

\bibliographystyle{plainnat}
\bibliography{aaai2026}

\clearpage
\appendix

\section*{Appendix Overview}

This appendix provides supporting derivations, implementation details,
runtime measurements, full numerical results, architecture-dependent
boundary checks, layerwise diagnostics, additional dataset checks, and
robustness experiments. Appendix~\nameref{app:repair_details} gives the details of
the fixed channelwise repair operator used to define RR.
Appendix~\nameref{app:allocation_details} gives the full greedy allocation
pseudocode. Appendix~\nameref{app:rr_derivation} provides an additional derivation
of the RR allocation objective and explains how the normalized
post-repair residual relates to the discrete budgeted allocation
problem. Appendix~\nameref{app:implementation} describes calibration, repair,
candidate sparsities, and baseline settings. Appendix~\nameref{app:runtime} reports
diagnostic and evaluation cost. Appendix~\nameref{app:full_results} reports paired
seed-level statistics, the full fine-grained ResNet18 transition sweep,
and representative achieved sparsities. Appendix~\nameref{app:layerwise} reports
the full layerwise comparison between RR and ERK.
Appendix~\nameref{app:architecture_results} summarizes the
architecture-dependent recoverability boundary. Appendix~\nameref{app:cts_detection} reports the full calibration-only transition detection results. Appendix~\nameref{app:local_joint_consistency} provides a local-to-joint
consistency check between the allocation-level RR objective and final
jointly-pruned recovery. 
Appendix~\nameref{app:vgg_results}
and Appendix~\nameref{app:resnet34_results} report the full VGG16-BN and ResNet34
results that support the boundary-shift analysis.
Appendix~\nameref{app:full_achieved_sparsities} reports full per-seed achieved
sparsities for VGG16-BN and ResNet34. Appendix~\nameref{app:additional_datasets}
reports transfer-style dataset checks. Finally, Appendix~\nameref{app:cal_sensitivity}
reports a calibration set size sensitivity study.

\section{Repair Operator Details}
\label{app:repair_details}

This appendix provides the implementation details of the fixed repair
operator used to define RR. The repair operator is not optimized during
allocation; it is held fixed and used both as a post-pruning correction
procedure and as a diagnostic instrument for measuring post-repair
residual distortion.

For a pruned convolutional layer, we estimate dense and pruned activation
statistics on the unlabeled calibration set. Let \(\sigma^2_{d,c}\) and
\(\sigma^2_{p,c}\) denote the dense and pruned activation variances of
output channel \(c\), respectively. A direct channelwise
variance-matching correction would multiply the corresponding output
filter by
\begin{equation}
\gamma_c^{\mathrm{direct}}
=
\sqrt{
\frac{\sigma^2_{d,c}+\epsilon}
     {\sigma^2_{p,c}+\epsilon}
}.
\label{eq:app_direct_scale}
\end{equation}
This correction restores the pruned activation variance toward the dense
reference variance. However, when pruning collapses a channel and
\(\sigma^2_{p,c}\) is close to zero, the direct ratio can become very
large and may amplify noise rather than restore useful signal.

We therefore apply a shrinkage-stabilized correction. We first define
the log-variance correction
\begin{equation}
r_c
=
\log(\sigma^2_{d,c}+\epsilon)
-
\log(\sigma^2_{p,c}+\epsilon).
\label{eq:app_log_variance_ratio}
\end{equation}
The correction is then shrunk toward zero using the reliability weight
\begin{equation}
\lambda_c
=
\frac{\sigma^2_{p,c}}
     {\sigma^2_{p,c}+\tau_\ell},
\label{eq:app_shrinkage_weight}
\end{equation}
where \(\tau_\ell\) is a layer-level shrinkage scale estimated from the
post-pruning channel variances on the calibration set. The applied
channel scale is
\begin{equation}
\gamma_c
=
\exp\!\left(\frac{1}{2}\lambda_c r_c\right).
\label{eq:app_repair_scale}
\end{equation}
Thus, channels whose post-pruning variance remains stable receive a
nearly full variance-matching correction, while channels whose variance
has collapsed receive an attenuated correction. The corresponding output
filter is rescaled by \(\gamma_c\).

After channelwise correction, we recalibrate BatchNorm running
statistics using unlabeled calibration batches, following standard
post-pruning practice. We refer to the combined pipeline as CR+BN. The
entire procedure uses only forward passes on calibration data and does
not use labels, gradients, or retraining. Because RR measures residual
distortion after this fixed repair operator, it should be interpreted as
a repair-operator-conditioned diagnostic rather than an intrinsic
property of a layer.

\section{Greedy Allocation Details}
\label{app:allocation_details}

This appendix gives the full pseudocode for the diagnostic-based greedy
allocation solver used in the main text. The same solver is used for all
calibration-based diagnostics, including raw activation shift,
unnormalized repair residual, and RR. This design keeps the comparison
focused on the diagnostic score rather than on differences in the
allocation procedure.

\begin{algorithm}[t]
\caption{Greedy diagnostic-based sparsity allocation}
\label{alg:app_allocation}
\begin{algorithmic}[1]
\Require Layers \(\{1,\ldots,L\}\), candidate sparsities
\(\mathcal{S}=\{s_1 < \cdots < s_K\}\), diagnostic score
\(g(\ell,s)\), target global sparsity \(S\), parameter counts
\(\{n_\ell\}_{\ell=1}^{L}\)
\State Initialize \(k_\ell \gets 1\) and \(s_\ell \gets s_1\) for all layers \(\ell\)
\State Compute
\[
S_{\mathrm{cur}}
\gets
\frac{\sum_{\ell=1}^{L} s_\ell n_\ell}
     {\sum_{\ell=1}^{L} n_\ell}
\]
\While{\(S_{\mathrm{cur}} < S\)}
  \State \(\mathcal{P} \gets \{\ell : k_\ell < K\}\)
  \State For each \(\ell \in \mathcal{P}\), compute
  \[
  q_\ell
  =
  \frac{
  g(\ell,s_{k_\ell+1})-g(\ell,s_{k_\ell})
  }{
  (s_{k_\ell+1}-s_{k_\ell}) n_\ell
  }.
  \]
  \State Select
  \[
  \ell^{*}
  \gets
  \arg\min_{\ell \in \mathcal{P}} q_\ell .
  \]
  \State \(k_{\ell^{*}} \gets k_{\ell^{*}}+1\)
  \State \(s_{\ell^{*}} \gets s_{k_{\ell^{*}}}\)
  \State Update
  \[
  S_{\mathrm{cur}}
  \gets
  \frac{\sum_{\ell=1}^{L} s_\ell n_\ell}
       {\sum_{\ell=1}^{L} n_\ell}
  \]
\EndWhile
\State \Return \(\{s_\ell\}_{\ell=1}^{L}\)
\end{algorithmic}
\end{algorithm}

The greedy step selects the candidate sparsity increment with the
smallest increase in diagnostic cost per additionally pruned parameter.
The normalization by \(n_\ell\) is important because layers can differ
substantially in parameter count. A fixed increase in diagnostic score is
less costly if it removes many more parameters.

The diagnostic curves are not required to be monotone over the candidate
grid. Negative increments are allowed and correspond to candidate
sparsity steps whose measured diagnostic cost decreases relative to the
previous grid point. We keep this behavior rather than imposing a
monotonicity constraint, since such a constraint would introduce an
additional modeling choice. By applying the same allocation solver to all
diagnostic scores, differences in recovered accuracy can be attributed to
the diagnostic signal itself.

The objective solved by the greedy procedure is only a diagnostic proxy.
The per-layer scores are measured using single-layer perturbations, while
the final sparse model prunes all layers simultaneously. We therefore do
not claim that the greedy allocation is globally optimal for final test
accuracy. Its role is to provide a controlled and reproducible way to
convert a local diagnostic score into a full sparsity allocation under a
fixed global budget.

\section{Additional Derivation of the RR Allocation Objective}
\label{app:rr_derivation}

This appendix gives a more detailed motivation for the diagnostic
objective used in Sec.~\nameref{sec:method}. The derivation is not intended as a
claim of global optimality. Rather, it explains why a post-repair
normalized residual is a natural local proxy for repair-aware sparsity
allocation.

\subsection{Local perturbation view}

Let \(F_\ell(x)\) denote the dense activation at layer \(\ell\), and let
\(F_{\ell,s}^{\mathrm{pruned}}(x)\) denote the activation obtained by
pruning only layer \(\ell\) to sparsity \(s\), keeping all other layers
dense. The raw local perturbation is
\[
\Delta_{\ell,s}^{\mathrm{raw}}(x)
=
F_{\ell,s}^{\mathrm{pruned}}(x)-F_\ell(x).
\]
A raw activation-shift diagnostic estimates the normalized magnitude of
this perturbation:
\[
D_\ell^{\mathrm{raw}}(s)
\approx
\mathbb{E}_{x\sim\mathcal{C}}
\left[
\|\Delta_{\ell,s}^{\mathrm{raw}}(x)\|^2_{\mathrm{norm}}
\right].
\]
This quantity measures how strongly pruning perturbs the representation,
but it does not measure whether the perturbation is compatible with the
repair operator.

Let \(\mathcal{T}\) denote the fixed channelwise repair operator. The
post-repair activation is
\[
F_{\ell,s}^{\mathrm{repair}}(x)
=
\mathcal{T}\!\left(F_{\ell,s}^{\mathrm{pruned}}(x)\right),
\]
and the residual perturbation is
\[
\Delta_{\ell,s}^{\mathrm{repair}}(x)
=
F_{\ell,s}^{\mathrm{repair}}(x)-F_\ell(x).
\]
The residual diagnostic estimates
\[
D_\ell^{\mathrm{repair}}(s)
\approx
\mathbb{E}_{x\sim\mathcal{C}}
\left[
\|\Delta_{\ell,s}^{\mathrm{repair}}(x)\|^2_{\mathrm{norm}}
\right].
\]
This residual is closer to the quantity that matters after repair, but
as an absolute score it still mixes the raw perturbation scale with the
repair failure rate.

\subsection{Residual fraction as repair compatibility}

A repair-aware allocation should distinguish two cases that can have the
same absolute residual. In the first case, pruning causes large raw
distortion but repair removes most of it. In the second case, pruning
causes only moderate raw distortion but repair removes little of it.
These cases have different implications for allocation: the first layer
is more compatible with the repair operator, while the second is more
repair-resistant.

RR separates these effects by forming the normalized residual
\[
R_\ell(s)
=
\frac{D_\ell^{\mathrm{repair}}(s)+\epsilon}
     {D_\ell^{\mathrm{raw}}(s)+\epsilon}.
\]
When \(D_\ell^{\mathrm{raw}}(s)\) is not close to zero, this ratio
approximates the fraction of the pruning-induced distortion that remains
after repair:
\[
R_\ell(s)
\approx
\frac{
\mathbb{E}_{x\sim\mathcal{C}}
[
\|\Delta_{\ell,s}^{\mathrm{repair}}(x)\|^2_{\mathrm{norm}}
]
}{
\mathbb{E}_{x\sim\mathcal{C}}
[
\|\Delta_{\ell,s}^{\mathrm{raw}}(x)\|^2_{\mathrm{norm}}
]
}.
\]
Thus \(R_\ell(s)\) is small when the repair operator removes most of the
local distortion, and large when the distortion persists. Values above
one indicate that the repair operator increases the measured discrepancy
relative to the raw pruned activation.

\subsection{Discrete budgeted allocation}

Given a local diagnostic score \(g(\ell,s)\), allocation over a discrete
candidate grid can be written as
\[
\min_{\{s_\ell \in \mathcal{S}\}_{\ell=1}^{L}}
\sum_{\ell=1}^{L} g(\ell,s_\ell)
\quad
\text{subject to}
\quad
\sum_{\ell=1}^{L} n_\ell s_\ell
\ge
S\sum_{\ell=1}^{L} n_\ell .
\]
This is a budgeted discrete allocation problem. The objective is
separable in the diagnostic proxy, while the sparsity constraint couples
the layer choices. Since \(g(\ell,s)\) is measured from local
single-layer perturbations, it does not include all cross-layer
interactions in the final simultaneously pruned network. We therefore
use the objective as a diagnostic allocation proxy, not as an exact
surrogate for final accuracy.

The greedy algorithm used in the main text chooses the next layer
promotion with the lowest marginal diagnostic cost per additionally
pruned parameter:
\[
q_\ell
=
\frac{
g(\ell,s_{k_\ell+1})-g(\ell,s_{k_\ell})
}{
(s_{k_\ell+1}-s_{k_\ell})n_\ell
}.
\]
This rule is analogous to selecting the cheapest local sparsity
increment under a parameter budget. It also ensures that layers with
different parameter counts are compared on the same scale: an increase
in diagnostic cost is penalized less if it removes substantially more
parameters.

\subsection{Why the same solver is used for all diagnostics}

The purpose of the experiments is to compare diagnostic signals rather
than optimization solvers. For this reason, raw activation shift,
post-repair residual, and RR are all passed to the same greedy allocation
procedure. If RR improves recovered accuracy under this shared solver,
the improvement is attributable to the normalized repairability signal
rather than to a different search procedure.

The diagnostic curves may be non-monotone on the discrete grid. This can
occur because the channelwise repair response changes with sparsity and
because the calibration estimate is finite-sample. We allow negative
marginal increments in the greedy procedure. Imposing monotonicity would
introduce an additional modeling choice and would make it harder to
separate the effect of the diagnostic from the effect of post-processing
the diagnostic curve.

\section{Implementation Details}
\label{app:implementation}

\paragraph{Calibration.}
All activation diagnostics use 128 unlabeled training images, collected
as two batches of 64. The same calibration examples are reused across
diagnostics within a seed to make comparisons controlled.

\paragraph{Repair and BatchNorm recalibration.}
After pruning, we apply channelwise variance-matching repair followed by
BatchNorm recalibration. BatchNorm running statistics are recalibrated
using 20 batches of 128 training images. No labels and no gradient
updates are used during repair.

\paragraph{Candidate sparsities.}
The diagnostic allocation rules use the candidate sparsity set
\[
\mathcal{S} =
\{0.70, 0.80, 0.85, 0.90, 0.925, 0.95, 0.975\}.
\]
Each layer is initialized at the lowest candidate sparsity. The greedy
allocation procedure repeatedly promotes one layer to its next candidate
sparsity according to the smallest incremental diagnostic cost per
additionally pruned parameter until the target global sparsity is
reached.

\paragraph{Architecture and training.}
The main mechanism experiments use ResNet18~\cite{he2016deep}
initialized from ImageNet-pretrained weights and fine-tuned on the
target dataset for five epochs with frozen lower layers. The first
convolutional layer is kept dense in all allocation methods. The
architecture-dependent boundary checks additionally evaluate VGG16-BN
and ResNet34 on CIFAR10 and CIFAR100.

\paragraph{ERK and LAMP settings.}
For ERK we use a minimum density floor of 0.025 and a maximum density cap
of 1.0. The density cap is important at high sparsity because small
projection layers can receive uncapped ERK densities above one, forcing
full retention under the standard capped implementation. For LAMP, we
compute layer-adaptive magnitude scores over the same repairable
convolutional layers and prune globally according to those normalized
scores. LAMP does not use calibration images.

\section{Runtime and Diagnostic Cost}
\label{app:runtime}

All diagnostics use 128 unlabeled calibration images and 20 BatchNorm
recalibration batches. On CIFAR100 ResNet18, the full allocation
comparison pipeline takes a median of 7.23 minutes per sparsity point
after the dense checkpoint is available. This timing includes global,
uniform, raw-shift, repair-residual, RR, and ERK evaluations, so the RR
diagnostic itself is only a subset of the reported time. The first run
for seed 0 takes 20.87 minutes because it includes dense model
fine-tuning; subsequent repeated sparsity points take about 7.2 minutes.

\begin{table}[t]
\centering
\small
\begin{tabular}{lc}
\toprule
Item & Value \\
\midrule
Calibration images & 128 \\
Candidate sparsities & 7 \\
Repairable convolutional layers & 19 \\
BN recalibration batches & 20 \\
Gradient updates during diagnosis & 0 \\
Median time per sparsity point & 7.23 min \\
Mean time per sparsity point & 8.72 min \\
First run with dense fine-tuning & 20.87 min \\
\bottomrule
\end{tabular}
\caption{Diagnostic and evaluation cost for the fine-grained transition
sweep on CIFAR100 ResNet18. Timing is measured for one sparsity point of
the full comparison pipeline, which includes all allocation baselines.}
\label{tab:app_diagnostic_cost}
\end{table}

\section{Full ResNet18 Results}
\label{app:full_results}

\subsection{Paired RR and ERK gaps}
\label{app:paired_gap}

For the comparison between RR and ERK, we report paired seed-level gaps.
For each seed, we first compute the accuracy difference between RR and
ERK, and then compute the mean and standard deviation across seeds. This
is the quantity used for the error bars in Fig.~\ref{fig:rr_erk_gap}. The
paired standard deviation differs from the propagated standard deviation
from independent single-method runs because RR and ERK are evaluated on
the same dense fine-tuned checkpoint per seed.

\begin{table}[t]
\centering
\small
\setlength{\tabcolsep}{6pt}
\begin{tabular}{lcc}
\toprule
Setting & RR \( - \) ERK mean & Paired std. \\
\midrule
CIFAR100 90\%   & \(-0.88\) & 1.72 \\
CIFAR100 92.5\% & \(-0.22\) & 0.83 \\
CIFAR100 95\%   & \(+2.48\) & 0.69 \\
CIFAR100 97.5\% & \(+0.00\) & 0.00 \\
\midrule
CIFAR10 90\%    & \(+0.23\) & 0.25 \\
CIFAR10 92.5\%  & \(+2.39\) & 0.97 \\
CIFAR10 95\%    & \(+6.18\) & 0.82 \\
CIFAR10 97.5\%  & \(+0.00\) & 0.00 \\
\bottomrule
\end{tabular}
\caption{Paired seed-level accuracy gaps after CR+BN between RR and ERK.
Positive values indicate that RR improves over ERK.}
\label{tab:app_paired_gap}
\end{table}

\subsection{Fine-grained transition table}
\label{app:fine_transition_full}

\begin{table*}[t]
\centering
\small
\setlength{\tabcolsep}{3.5pt}
\begin{tabular}{lccccccc}
\toprule
Target sparsity & Global & Uniform & Raw Shift & Repair Residual & RR & LAMP & ERK \\
\midrule
93.5\% & \(5.61 \pm 0.76\) & \(7.21 \pm 0.45\) & \(17.01 \pm 0.27\) & \(16.89 \pm 0.28\) & \(18.54 \pm 0.57\) & -- & \(18.19 \pm 0.25\) \\
94.0\% & \(4.69 \pm 1.04\) & \(6.04 \pm 0.30\) & \(13.27 \pm 0.06\) & \(15.13 \pm 0.53\) & \(17.68 \pm 0.35\) & \(17.43 \pm 0.46\) & \(14.56 \pm 0.79\) \\
94.5\% & \(4.40 \pm 0.89\) & \(4.38 \pm 0.39\) & \(11.31 \pm 0.24\) & \(12.07 \pm 0.19\) & \(14.20 \pm 0.55\) & \(14.19 \pm 0.57\) & \(11.61 \pm 0.32\) \\
95.0\% & \(3.80 \pm 0.57\) & \(3.54 \pm 0.15\) & \(10.00 \pm 0.06\) & \(10.06 \pm 0.30\) & \(11.88 \pm 0.49\) & \(10.76 \pm 0.15\) & \(9.38 \pm 0.15\) \\
95.5\% & \(3.42 \pm 0.39\) & \(2.68 \pm 0.21\) & \(6.93 \pm 0.53\) & \(7.66 \pm 0.54\) & \(9.55 \pm 0.63\) & \(7.93 \pm 0.30\) & \(7.23 \pm 0.40\) \\
96.0\% & \(2.86 \pm 0.47\) & \(1.96 \pm 0.32\) & \(5.16 \pm 0.31\) & \(5.23 \pm 0.36\) & \(5.56 \pm 0.18\) & -- & \(5.31 \pm 0.22\) \\
\bottomrule
\end{tabular}
\caption{Full fine-grained transition sweep on CIFAR100 ResNet18.
Results report recovered accuracy after CR+BN, mean \(\pm\) standard
deviation over three seeds. LAMP is evaluated on the central transition
band from 94.0\% to 95.5\%.}
\label{tab:app_fine_transition_full}
\end{table*}

\subsection{Representative achieved sparsities}
\label{app:actual_sparsity}

The allocation rules occasionally produce slightly different achieved
global sparsities because the allocation is selected from a discrete set
of candidate layer sparsities. Table~\ref{tab:app_actual_sparsity} reports
representative achieved sparsities for key ResNet18 settings. The
differences are small and do not explain the observed accuracy gains.

\begin{table}[t]
\centering
\small
\setlength{\tabcolsep}{4pt}
\begin{tabular}{lcccc}
\toprule
Setting & Raw & Repair Res. & RR & ERK \\
\midrule
CIFAR10 92.5\% & 92.65 & 92.65 & 92.61 & 92.42 \\
CIFAR10 95\%   & 95.03 & 95.03 & 94.96 & 94.92 \\
CIFAR10 97.5\% & 97.42 & 97.42 & 97.42 & 97.42 \\
CIFAR100 97.5\% & 97.42 & 97.42 & 97.42 & 97.42 \\
\bottomrule
\end{tabular}
\caption{Achieved global sparsities (\%) for the four diagnostic
allocation rules. Differences are small; RR does not obtain its CIFAR10
gains by using a lower effective sparsity.}
\label{tab:app_actual_sparsity}
\end{table}

\section{Layerwise Allocation Diagnostics}
\label{app:layerwise}

We report the layerwise sparsity comparison between RR and ERK on
CIFAR100 at 95\% sparsity. This setting corresponds to
Fig.~\ref{fig:rr_erk_deviation} of the main paper. The two allocations are
highly correlated, but the deviations are stable across seeds: positive
deviations indicate layers where RR assigns higher sparsity than ERK,
while negative deviations indicate layers it protects more strongly.

\begin{table*}[t]
\centering
\small
\begin{tabular}{lccc}
\toprule
Layer & ERK sparsity & RR sparsity & RR \( - \) ERK \\
\midrule
\texttt{layer1.0.conv1}        & 0.764 & 0.800 & \(+0.036\) \\
\texttt{layer1.0.conv2}        & 0.764 & 0.850 & \(+0.086\) \\
\texttt{layer1.1.conv1}        & 0.764 & 0.700 & \(-0.064\) \\
\texttt{layer1.1.conv2}        & 0.764 & 0.850 & \(+0.086\) \\
\texttt{layer2.0.conv1}        & 0.826 & 0.700 & \(-0.126\) \\
\texttt{layer2.0.conv2}        & 0.885 & 0.850 & \(-0.035\) \\
\texttt{layer2.0.downsample.0} & 0.000 & 0.700 & \(+0.700\) \\
\texttt{layer2.1.conv1}        & 0.885 & 0.850 & \(-0.035\) \\
\texttt{layer2.1.conv2}        & 0.885 & 0.850 & \(-0.035\) \\
\texttt{layer3.0.conv1}        & 0.914 & 0.850 & \(-0.064\) \\
\texttt{layer3.0.conv2}        & 0.943 & 0.950 & \(+0.007\) \\
\texttt{layer3.0.downsample.0} & 0.237 & 0.733 & \(+0.497\) \\
\texttt{layer3.1.conv1}        & 0.943 & 0.900 & \(-0.043\) \\
\texttt{layer3.1.conv2}        & 0.943 & 0.950 & \(+0.007\) \\
\texttt{layer4.0.conv1}        & 0.957 & 0.950 & \(-0.007\) \\
\texttt{layer4.0.conv2}        & 0.972 & 0.975 & \(+0.003\) \\
\texttt{layer4.0.downsample.0} & 0.619 & 0.817 & \(+0.197\) \\
\texttt{layer4.1.conv1}        & 0.972 & 0.975 & \(+0.003\) \\
\texttt{layer4.1.conv2}        & 0.972 & 0.975 & \(+0.003\) \\
\bottomrule
\end{tabular}
\caption{Layerwise sparsity comparison between ERK and RR on CIFAR100 at
95\% sparsity. The largest deviations occur in projection layers, where
ERK assigns unusually low sparsity due to its density cap.}
\label{tab:app_layerwise_delta}
\end{table*}

\paragraph{ERK density cap diagnostic.}
The largest deviation in Table~\ref{tab:app_layerwise_delta} occurs in
\texttt{layer2.0.downsample.0}. Under the standard capped ERK rule, the
uncapped density of this layer exceeds one and is clipped, forcing the
layer to be retained at full density. RR instead assigns 70\% sparsity
to this layer, suggesting that its pruning-induced distortion is
recoverable under the CR+BN pipeline.

\begin{table}[t]
\centering
\small
\setlength{\tabcolsep}{5pt}
\begin{tabular}{lccc}
\toprule
Layer & Uncapped density & ERK sparsity & RR sparsity \\
\midrule
\texttt{layer2.0.downsample.0} & 1.534 & 0.000 & 0.700 \\
\texttt{layer3.0.downsample.0} & 0.763 & 0.237 & 0.733 \\
\texttt{layer4.0.downsample.0} & 0.381 & 0.619 & 0.817 \\
\bottomrule
\end{tabular}
\caption{Projection-layer diagnostic on CIFAR100 at 95\% sparsity. ERK
fully retains \texttt{layer2.0.downsample.0} because its uncapped
density exceeds one. RR assigns substantial sparsity to all three
projection layers while improving recovered accuracy.}
\label{tab:app_erk_cap}
\end{table}

\section{Architecture-Dependent Boundary Results}
\label{app:architecture_results}

The main text summarizes the architecture-dependent recoverability
boundary using a compact table. Here we provide the full VGG16-BN and
ResNet34 results. These experiments are not used to claim that RR
dominates across all architectures and sparsities. Instead, they support
the interpretation that the useful region of RR shifts with the
architecture's collapse boundary: VGG16-BN reaches this boundary earlier,
whereas ResNet34 reaches it later.

Across both architectures, achieved sparsities are closely matched under
the same target sparsity. Therefore, the observed gains near the boundary
are not explained by relaxed pruning constraints.

\section{Full Calibration-only Transition Detection Results}
\label{app:cts_detection}

This appendix reports the full per-seed outputs of the calibration-only transition detector introduced in Sec.~\nameref{sec:calibration_transition_detector}. 
The detector uses only RR diagnostic curves and does not use test accuracy. 
For each architecture--dataset--seed setting, we report two quantities.

First, the broad detected band is a less selective contiguous sparsity region where the calibration transition score is elevated. 
It is useful for visualizing the wider region in which repairability disagreement begins to appear.

Second, the high-confidence core is the stricter contiguous interval selected by the fixed threshold in Sec.~\nameref{sec:calibration_transition_detector}. 
This is the quantity summarized in the main text. 
It is intended to identify the central region where repair-aware allocation is expected to matter most.

Table~\ref{tab:cts_detection_per_seed_band} shows the full per-seed detected bands and high-confidence cores for ResNet18, VGG16-BN, and ResNet34. 
We omit ResNet50 from the main analysis to keep the architecture-dependent transition story focused on the models reported in the main paper.

On CIFAR100 ResNet18, all three seeds select $95.0$--$95.5\%$ as the high-confidence core. 
This agrees with the upper part of the accuracy-based transition band, where RR surpasses LAMP. 
On CIFAR10 ResNet18, the broader detector does not form a stable contiguous band, but the high-confidence detector consistently selects $97.0\%$, indicating a late high-stress region on the easier dataset.

For VGG16-BN, the detector shifts the transition later than in the previous broad summary. 
On CIFAR10, two seeds select $93.5$--$95.5\%$ as the high-confidence core, while one seed selects $92.0$--$93.5\%$. 
On CIFAR100, all three seeds select $94.0$--$95.5\%$. 
This supports the interpretation that VGG16-BN enters a repairability-sensitive region near the collapse boundary, rather than at a single fixed sparsity point.

For ResNet34, the detected core moves to the extreme high-sparsity region. 
On CIFAR10, all three seeds select $97.5\%$ as the high-confidence core. 
On CIFAR100, the broad detected band is consistently $96.0$--$98.0\%$, with high-confidence cores at $97.0\%$ or $97.5\%$. 
This is consistent with the main architecture-dependent boundary interpretation: ResNet34 remains stable under standard allocation rules until more extreme sparsity, and RR becomes most informative only near the later recoverability boundary.

Overall, these per-seed results support the use of CTS as a calibration-only indicator of RR-useful transition regions. 
The detected cores align well with the sparsity ranges where RR becomes strongest or most competitive in recovered accuracy, while still being interpreted as diagnostic indicators rather than exact predictors of final accuracy.

\begin{table*}[t]
\centering
\scriptsize
\setlength{\tabcolsep}{4pt}
\caption{Per-seed calibration-only transition detection results. 
The detector uses only RR diagnostic curves and does not use test accuracy. 
We report the broad detected band and the high-confidence transition core.}
\label{tab:cts_detection_per_seed_band}
\begin{tabular}{lllcc}
\toprule
Architecture & Dataset & Seed & Broad detected band & High-confidence core \\
\midrule
ResNet18 & CIFAR100 & 0 & 94.0--96.0\% & 95.0--95.5\% \\
ResNet18 & CIFAR100 & 1 & 94.5--95.5\% & 95.0--95.5\% \\
ResNet18 & CIFAR100 & 2 & 94.5--95.5\% & 95.0--95.5\% \\
\midrule
ResNet18 & CIFAR10 & 0 & not detected & 97.0\% \\
ResNet18 & CIFAR10 & 1 & not detected & 97.0\% \\
ResNet18 & CIFAR10 & 2 & not detected & 97.0\% \\
\midrule
VGG16-BN & CIFAR10 & 0 & 93.0--96.0\% & 93.5--95.5\% \\
VGG16-BN & CIFAR10 & 1 & 93.0--96.0\% & 93.5--95.5\% \\
VGG16-BN & CIFAR10 & 2 & 93.0--96.0\% & 92.0--93.5\% \\
\midrule
VGG16-BN & CIFAR100 & 0 & 93.0--95.5\% & 94.0--95.5\% \\
VGG16-BN & CIFAR100 & 1 & 93.0--96.0\% & 94.0--95.5\% \\
VGG16-BN & CIFAR100 & 2 & 93.0--96.0\% & 94.0--95.5\% \\
\midrule
ResNet34 & CIFAR10 & 0 & 96.5--97.5\% & 97.5\% \\
ResNet34 & CIFAR10 & 1 & 96.5--97.5\% & 97.5\% \\
ResNet34 & CIFAR10 & 2 & 96.5--97.5\% & 97.5\% \\
\midrule
ResNet34 & CIFAR100 & 0 & 96.0--98.0\% & 97.0\% \\
ResNet34 & CIFAR100 & 1 & 96.0--98.0\% & 97.5\% \\
ResNet34 & CIFAR100 & 2 & 96.0--98.0\% & 97.5\% \\
\bottomrule
\end{tabular}
\end{table*}
\section{Local Diagnostic and Joint-Pruning Consistency}
\label{app:local_joint_consistency}

RR uses local single-layer perturbations to estimate repairability, while the final sparse model prunes all layers jointly. 
This creates a potential mismatch: the diagnostic objective does not explicitly model cross-layer interactions in the final jointly-pruned network. 
To assess whether the local diagnostic is still useful as an allocation-level proxy, we compare the RR diagnostic objective of an allocation with its final recovered accuracy after joint pruning and CR+BN repair.

For an allocation $A$, we compute
\[
J_{\mathrm{RR}}(A)
=
\frac{1}{L}
\sum_{\ell=1}^{L}
R_{\ell}(s_{\ell}),
\]
where $s_{\ell}$ is the sparsity assigned to layer $\ell$ by allocation $A$, and $R_{\ell}(s_{\ell})$ is obtained by recomputing the exact single-layer RR diagnostic at the actual layer sparsity used by that allocation. 
No interpolation is used in this strict version. 
We then compare $-J_{\mathrm{RR}}(A)$ with the final CR+BN recovered accuracy of the jointly-pruned model. 
The negative sign is used because lower $J_{\mathrm{RR}}$ indicates less repair-resistant distortion and should therefore correspond to higher recovered accuracy.

We perform this analysis on CIFAR100 ResNet18 in the transition band at 94.5\%, 95.0\%, and 95.5\% sparsity. 
For each sparsity and seed, we compute rank correlation across seven allocation rules: Global, Uniform, ERK, LAMP, Raw Shift, Repair Residual, and RR. 
This experiment is not intended to show that the local objective exactly predicts final accuracy. 
Rather, it tests whether the local repairability diagnostic provides a useful ranking signal for final joint-pruning recovery.

\begin{table}[t]
\centering
\small
\caption{Consistency between the local RR diagnostic objective and final jointly-pruned recovery on CIFAR100 ResNet18. 
We report rank correlation between $-J_{\mathrm{RR}}(A)$ and final CR+BN recovered accuracy across allocation rules, averaged over three seeds.}
\label{tab:local_joint_consistency}
\begin{tabular}{lcc}
\toprule
Sparsity & Spearman $\rho$ & Kendall $\tau$ \\
\midrule
94.5\% & $0.131 \pm 0.074$ & $0.111 \pm 0.055$ \\
95.0\% & $0.312 \pm 0.120$ & $0.260 \pm 0.149$ \\
95.5\% & $0.488 \pm 0.055$ & $0.397 \pm 0.055$ \\
\bottomrule
\end{tabular}
\end{table}

Table~\ref{tab:local_joint_consistency} shows that the correlations are positive at all three sparsity levels. 
The correlation is weak at 94.5\%, becomes stronger at 95.0\%, and is strongest at 95.5\%. 
This trend is consistent with the main observation that RR becomes more informative as the model moves deeper into the recoverability transition band. 
The result supports the use of the local RR objective as a practical allocation-level proxy.

At the same time, the correlations are moderate rather than perfect. 
This is expected because the local diagnostic does not explicitly model all cross-layer interactions in the final jointly-pruned model. 
Therefore, this analysis should not be interpreted as showing that $J_{\mathrm{RR}}(A)$ exactly predicts final recovered accuracy. 
Instead, it shows that allocations with lower local repair-resistant distortion tend to achieve better CR+BN recovery after joint pruning, while still leaving room for mismatch caused by joint pruning effects.

\section{VGG16-BN Boundary Results}
\label{app:vgg_results}

VGG16-BN is a plain feed-forward CNN and is more vulnerable to severe
degradation under high sparsity than residual networks. This setting
therefore tests whether repair-aware allocation becomes useful near an
earlier collapse boundary. Table~\ref{tab:app_vgg_main} reports the full
accuracy results. RR is weak at moderate sparsity but becomes effective
near the high-sparsity collapse boundary, especially on CIFAR10 at
92.5--95\% and on CIFAR100 at 95\%.

\begin{table*}[t]
\centering
\small
\setlength{\tabcolsep}{4pt}
\begin{tabular}{llccccccc}
\toprule
Dataset & Sparsity & Global & Uniform & ERK & LAMP & Raw Shift & Repair Residual & RR \\
\midrule
CIFAR10 & 85\% & \textbf{56.65$\pm$1.60} & 47.96$\pm$3.16 & 53.97$\pm$1.60 & 54.94$\pm$1.12 & 51.30$\pm$0.60 & 47.74$\pm$0.79 & 32.07$\pm$1.82 \\
CIFAR10 & 90\% & \textbf{31.13$\pm$0.91} & 27.71$\pm$1.18 & 29.64$\pm$1.58 & 25.75$\pm$3.44 & 26.55$\pm$0.40 & 19.75$\pm$2.30 & 25.82$\pm$1.10 \\
CIFAR10 & 92.5\% & 14.46$\pm$1.35 & 17.09$\pm$0.22 & 13.85$\pm$1.22 & 12.61$\pm$1.36 & 12.97$\pm$0.60 & \textbf{18.81$\pm$1.09} & 17.63$\pm$0.71 \\
CIFAR10 & 95\% & 13.05$\pm$0.97 & 10.13$\pm$0.08 & 10.54$\pm$0.32 & 10.11$\pm$0.03 & 10.03$\pm$0.05 & 10.11$\pm$0.10 & \textbf{14.82$\pm$0.84} \\
\midrule
CIFAR100 & 85\% & 21.69$\pm$0.85 & 12.59$\pm$0.92 & 22.69$\pm$0.65 & \textbf{23.11$\pm$0.90} & 17.95$\pm$0.37 & 16.05$\pm$1.12 & 10.43$\pm$0.72 \\
CIFAR100 & 90\% & 6.00$\pm$0.78 & 4.66$\pm$0.37 & 6.04$\pm$0.33 & 5.27$\pm$0.27 & 4.57$\pm$0.25 & 2.84$\pm$0.27 & \textbf{6.07$\pm$0.57} \\
CIFAR100 & 92.5\% & \textbf{3.08$\pm$0.26} & 2.66$\pm$0.10 & 3.02$\pm$0.19 & 2.20$\pm$0.30 & 2.30$\pm$0.11 & 2.63$\pm$0.41 & 2.50$\pm$0.37 \\
CIFAR100 & 95\% & 1.38$\pm$0.12 & 1.09$\pm$0.17 & 1.31$\pm$0.14 & 1.02$\pm$0.04 & 1.14$\pm$0.04 & 1.14$\pm$0.06 & \textbf{1.75$\pm$0.13} \\
\bottomrule
\end{tabular}
\caption{VGG16-BN results under CR+BN repair. We report mean top-1
accuracy over three seeds. RR is not uniformly superior at moderate
sparsity, but becomes effective near the high-sparsity collapse
boundary.}
\label{tab:app_vgg_main}
\end{table*}

Table~\ref{tab:app_vgg_rr_gaps} reports RR's gains over strong allocation
baselines. The gains are negative at 85\%, become mixed around 90\%, and
become positive near the highest sparsity levels. This supports the
interpretation that RR is useful near the collapse boundary rather than
uniformly across sparsity levels.

\begin{table}[t]
\centering
\small
\setlength{\tabcolsep}{4pt}
\begin{tabular}{llccc}
\toprule
Dataset & Sparsity & RR \( - \) ERK & RR \( - \) LAMP & RR \( - \) Raw \\
\midrule
CIFAR10 & 85\%   & \(-21.90 \pm 3.41\) & \(-22.87 \pm 2.91\) & \(-19.23 \pm 2.41\) \\
CIFAR10 & 90\%   & \(-3.82 \pm 1.03\)  & \(0.07 \pm 2.73\)   & \(-0.73 \pm 0.84\) \\
CIFAR10 & 92.5\% & \(3.78 \pm 1.92\)   & \(5.02 \pm 1.95\)   & \(4.66 \pm 1.25\) \\
CIFAR10 & 95\%   & \textbf{\(4.28 \pm 0.86\)} & \textbf{\(4.71 \pm 0.87\)} & \textbf{\(4.79 \pm 0.81\)} \\
\midrule
CIFAR100 & 85\%   & \(-12.27 \pm 1.29\) & \(-12.68 \pm 1.34\) & \(-7.52 \pm 1.08\) \\
CIFAR100 & 90\%   & \(0.03 \pm 0.76\)   & \(0.80 \pm 0.73\)   & \(1.50 \pm 0.82\) \\
CIFAR100 & 92.5\% & \(-0.53 \pm 0.35\)  & \(0.30 \pm 0.45\)   & \(0.19 \pm 0.28\) \\
CIFAR100 & 95\%   & \textbf{\(0.44 \pm 0.26\)} & \textbf{\(0.74 \pm 0.12\)} & \textbf{\(0.61 \pm 0.14\)} \\
\bottomrule
\end{tabular}
\caption{Accuracy gains of RR over strong allocation baselines on
VGG16-BN. Gains become positive mainly near the high-sparsity collapse
boundary.}
\label{tab:app_vgg_rr_gaps}
\end{table}

\section{ResNet34 Boundary Results}
\label{app:resnet34_results}

ResNet34 provides a complementary architecture check. Because residual
connections stabilize feature propagation under pruning, the collapse
boundary appears later than in VGG16-BN. Table~\ref{tab:app_resnet34_main}
reports the full accuracy results. RR does not dominate in the
90--95\% range, where ERK and LAMP remain strong. At 97.5\%, however,
RR becomes the best-performing allocation rule on both CIFAR10 and
CIFAR100.

\begin{table*}[t]
\centering
\small
\setlength{\tabcolsep}{4pt}
\begin{tabular}{llccccccc}
\toprule
Dataset & Sparsity & Global & Uniform & ERK & LAMP & Raw Shift & Repair Residual & RR \\
\midrule
CIFAR10 & 90\% & 58.26$\pm$1.74 & 55.61$\pm$0.96 & 78.29$\pm$0.17 & \textbf{79.08$\pm$0.36} & 76.79$\pm$0.32 & 76.82$\pm$0.09 & 78.70$\pm$0.28 \\
CIFAR10 & 92.5\% & 39.80$\pm$1.04 & 41.55$\pm$1.70 & 66.05$\pm$0.56 & \textbf{69.03$\pm$0.58} & 59.38$\pm$0.96 & 63.51$\pm$4.35 & 65.84$\pm$1.32 \\
CIFAR10 & 95\% & 23.07$\pm$1.47 & 29.87$\pm$1.51 & 45.34$\pm$1.75 & \textbf{49.24$\pm$2.32} & 39.47$\pm$2.62 & 40.84$\pm$0.99 & 42.07$\pm$2.07 \\
CIFAR10 & 97.5\% & 15.07$\pm$1.58 & 13.63$\pm$0.83 & 22.80$\pm$4.48 & 23.45$\pm$3.61 & 15.81$\pm$1.05 & 20.26$\pm$1.32 & \textbf{24.09$\pm$1.15} \\
\midrule
CIFAR100 & 90\% & 7.73$\pm$0.36 & 22.25$\pm$1.56 & \textbf{40.04$\pm$0.30} & 39.52$\pm$0.64 & 36.22$\pm$0.31 & 36.01$\pm$0.31 & 36.83$\pm$2.81 \\
CIFAR100 & 92.5\% & 3.45$\pm$0.44 & 10.54$\pm$0.76 & 24.56$\pm$0.25 & \textbf{25.33$\pm$0.13} & 15.93$\pm$0.46 & 18.95$\pm$0.33 & 20.37$\pm$0.29 \\
CIFAR100 & 95\% & 2.45$\pm$0.25 & 4.58$\pm$0.36 & \textbf{10.66$\pm$1.34} & 10.40$\pm$0.76 & 8.93$\pm$0.83 & 8.52$\pm$0.79 & 7.70$\pm$0.73 \\
CIFAR100 & 97.5\% & 1.79$\pm$0.21 & 1.69$\pm$0.14 & 2.51$\pm$0.76 & 2.87$\pm$0.34 & 2.56$\pm$0.38 & 2.80$\pm$0.56 & \textbf{3.28$\pm$0.49} \\
\bottomrule
\end{tabular}
\caption{ResNet34 results under CR+BN repair. We report mean top-1
accuracy over three seeds. RR is not strongest in the 90--95\% range,
but becomes the best-performing method at the extreme 97.5\% sparsity
level on both CIFAR10 and CIFAR100.}
\label{tab:app_resnet34_main}
\end{table*}

\begin{table*}[t]
\centering
\small
\setlength{\tabcolsep}{4pt}
\begin{tabular}{llccc}
\toprule
Dataset & Sparsity & RR \( - \) ERK & RR \( - \) LAMP & RR \( - \) Raw \\
\midrule
CIFAR10 & 90\%   & \(0.41 \pm 0.46\) & \(-0.38 \pm 0.51\) & \(1.91 \pm 0.40\) \\
CIFAR10 & 92.5\% & \(-0.22 \pm 0.77\) & \(-3.19 \pm 1.02\) & \(6.45 \pm 0.39\) \\
CIFAR10 & 95\%   & \(-3.27 \pm 3.60\) & \(-7.17 \pm 3.85\) & \(2.60 \pm 3.97\) \\
CIFAR10 & 97.5\% & \textbf{\(1.28 \pm 5.62\)} & \textbf{\(0.63 \pm 4.73\)} & \textbf{\(8.27 \pm 1.90\)} \\
\midrule
CIFAR100 & 90\%   & \(-3.21 \pm 2.82\) & \(-2.70 \pm 2.80\) & \(0.61 \pm 3.07\) \\
CIFAR100 & 92.5\% & \(-4.19 \pm 0.10\) & \(-4.96 \pm 0.25\) & \(4.44 \pm 0.17\) \\
CIFAR100 & 95\%   & \(-2.96 \pm 1.09\) & \(-2.70 \pm 0.83\) & \(-1.23 \pm 0.48\) \\
CIFAR100 & 97.5\% & \textbf{\(0.77 \pm 0.34\)} & \textbf{\(0.41 \pm 0.16\)} & \textbf{\(0.72 \pm 0.28\)} \\
\bottomrule
\end{tabular}
\caption{Accuracy gains of RR over strong allocation baselines on
ResNet34. RR does not dominate ERK or LAMP in the 90--95\% range, but
outperforms them at the extreme 97.5\% sparsity level.}
\label{tab:app_resnet34_rr_gaps}
\end{table*}

\section{Full Achieved Sparsity Tables}
\label{app:full_achieved_sparsities}

This section reports full per-seed achieved sparsities for VGG16-BN and
ResNet34. The achieved sparsities are closely matched across allocation
methods under the same target sparsity. Therefore, the improvements of
RR near the recoverability boundary are not caused by using a less sparse
model.

\begin{table*}[t]
\centering
\small
\setlength{\tabcolsep}{3.5pt}
\begin{tabular}{lllccccccc}
\toprule
Dataset & Target & Seed & Global & Uniform & ERK & LAMP & Raw Shift & Repair Residual & RR \\
\midrule
CIFAR10 & 85\% & 0 & 85.00 & 84.99 & 84.99 & 84.99 & 85.53 & 85.13 & 85.73 \\
CIFAR10 & 85\% & 1 & 85.00 & 84.99 & 84.99 & 84.99 & 85.53 & 85.13 & 85.73 \\
CIFAR10 & 85\% & 2 & 85.00 & 84.99 & 84.99 & 84.99 & 85.53 & 85.13 & 85.73 \\
CIFAR10 & 90\% & 0 & 90.00 & 89.99 & 89.99 & 89.99 & 90.74 & 90.04 & 90.00 \\
CIFAR10 & 90\% & 1 & 90.00 & 89.99 & 89.99 & 89.99 & 90.74 & 90.74 & 90.00 \\
CIFAR10 & 90\% & 2 & 90.00 & 89.99 & 89.99 & 89.99 & 90.74 & 90.04 & 90.00 \\
CIFAR10 & 92.5\% & 0 & 92.50 & 92.49 & 92.49 & 92.49 & 92.85 & 92.85 & 92.82 \\
CIFAR10 & 92.5\% & 1 & 92.50 & 92.49 & 92.49 & 92.49 & 92.85 & 92.85 & 92.82 \\
CIFAR10 & 92.5\% & 2 & 92.50 & 92.49 & 92.49 & 92.49 & 92.85 & 92.85 & 92.82 \\
CIFAR10 & 95\% & 0 & 95.00 & 94.99 & 94.99 & 94.99 & 95.05 & 95.25 & 95.01 \\
CIFAR10 & 95\% & 1 & 95.00 & 94.99 & 94.99 & 94.99 & 95.05 & 95.25 & 95.01 \\
CIFAR10 & 95\% & 2 & 95.00 & 94.99 & 94.99 & 94.99 & 95.05 & 95.25 & 95.01 \\
\midrule
CIFAR100 & 85\% & 0 & 85.00 & 84.99 & 84.99 & 84.99 & 85.53 & 85.13 & 85.73 \\
CIFAR100 & 85\% & 1 & 85.00 & 84.99 & 84.99 & 84.99 & 85.53 & 85.13 & 85.73 \\
CIFAR100 & 85\% & 2 & 85.00 & 84.99 & 84.99 & 84.99 & 85.53 & 85.13 & 85.73 \\
CIFAR100 & 90\% & 0 & 90.00 & 89.99 & 89.99 & 89.99 & 90.74 & 90.04 & 90.00 \\
CIFAR100 & 90\% & 1 & 90.00 & 89.99 & 89.99 & 89.99 & 90.74 & 90.04 & 90.00 \\
CIFAR100 & 90\% & 2 & 90.00 & 89.99 & 89.99 & 89.99 & 90.74 & 90.04 & 90.00 \\
CIFAR100 & 92.5\% & 0 & 92.50 & 92.49 & 92.49 & 92.49 & 92.85 & 92.85 & 92.82 \\
CIFAR100 & 92.5\% & 1 & 92.50 & 92.49 & 92.49 & 92.49 & 92.85 & 92.85 & 92.82 \\
CIFAR100 & 92.5\% & 2 & 92.50 & 92.49 & 92.49 & 92.49 & 92.85 & 92.65 & 92.82 \\
CIFAR100 & 95\% & 0 & 95.00 & 94.99 & 94.99 & 94.99 & 95.05 & 95.25 & 95.01 \\
CIFAR100 & 95\% & 1 & 95.00 & 94.99 & 94.99 & 94.99 & 95.05 & 95.25 & 95.01 \\
CIFAR100 & 95\% & 2 & 95.00 & 94.99 & 94.99 & 94.99 & 95.05 & 95.25 & 95.01 \\
\bottomrule
\end{tabular}
\caption{Actual sparsity (\%) on VGG16-BN. All allocation methods
achieve comparable actual sparsity under the same target sparsity.}
\label{tab:app_vgg_actual_sparsity_full}
\end{table*}

\begin{table*}[t]
\centering
\small
\setlength{\tabcolsep}{3.5pt}
\begin{tabular}{lllccccccc}
\toprule
Dataset & Target & Seed & Global & Uniform & ERK & LAMP & Raw Shift & Repair Residual & RR \\
\midrule
CIFAR10 & 90\% & 0 & 90.00 & 89.96 & 89.96 & 89.96 & 90.04 & 90.04 & 89.97 \\
CIFAR10 & 90\% & 1 & 90.00 & 89.96 & 89.96 & 89.96 & 90.04 & 90.04 & 89.97 \\
CIFAR10 & 90\% & 2 & 90.00 & 89.96 & 89.96 & 89.96 & 90.04 & 90.04 & 89.97 \\
CIFAR10 & 92.5\% & 0 & 92.50 & 92.46 & 92.46 & 92.46 & 92.50 & 92.50 & 92.67 \\
CIFAR10 & 92.5\% & 1 & 92.50 & 92.46 & 92.46 & 92.46 & 92.50 & 92.57 & 92.46 \\
CIFAR10 & 92.5\% & 2 & 92.50 & 92.46 & 92.46 & 92.46 & 92.50 & 92.50 & 92.68 \\
CIFAR10 & 95\% & 0 & 95.00 & 94.96 & 94.96 & 94.96 & 95.00 & 94.99 & 95.09 \\
CIFAR10 & 95\% & 1 & 95.00 & 94.96 & 94.96 & 94.96 & 95.00 & 94.99 & 95.02 \\
CIFAR10 & 95\% & 2 & 95.00 & 94.96 & 94.96 & 94.96 & 95.00 & 94.99 & 94.97 \\
CIFAR10 & 97.5\% & 0 & 97.50 & 97.46 & 97.46 & 97.46 & 97.47 & 97.47 & 97.47 \\
CIFAR10 & 97.5\% & 1 & 97.50 & 97.46 & 97.46 & 97.46 & 97.47 & 97.47 & 97.46 \\
CIFAR10 & 97.5\% & 2 & 97.50 & 97.46 & 97.46 & 97.46 & 97.47 & 97.47 & 97.46 \\
\midrule
CIFAR100 & 90\% & 0 & 90.00 & 89.96 & 89.96 & 89.96 & 90.04 & 90.04 & 89.97 \\
CIFAR100 & 90\% & 1 & 90.00 & 89.96 & 89.96 & 89.96 & 90.04 & 90.04 & 89.97 \\
CIFAR100 & 90\% & 2 & 90.00 & 89.96 & 89.96 & 89.96 & 90.04 & 90.04 & 90.18 \\
CIFAR100 & 92.5\% & 0 & 92.50 & 92.46 & 92.46 & 92.46 & 92.50 & 92.50 & 92.47 \\
CIFAR100 & 92.5\% & 1 & 92.50 & 92.46 & 92.46 & 92.46 & 92.50 & 92.50 & 92.46 \\
CIFAR100 & 92.5\% & 2 & 92.50 & 92.46 & 92.46 & 92.46 & 92.50 & 92.50 & 92.46 \\
CIFAR100 & 95\% & 0 & 95.00 & 94.96 & 94.96 & 94.96 & 95.00 & 95.00 & 95.01 \\
CIFAR100 & 95\% & 1 & 95.00 & 94.96 & 94.96 & 94.96 & 95.00 & 94.99 & 95.05 \\
CIFAR100 & 95\% & 2 & 95.00 & 94.96 & 94.96 & 94.96 & 95.00 & 95.00 & 95.03 \\
CIFAR100 & 97.5\% & 0 & 97.50 & 97.46 & 97.46 & 97.46 & 97.47 & 97.47 & 97.51 \\
CIFAR100 & 97.5\% & 1 & 97.50 & 97.46 & 97.46 & 97.46 & 97.47 & 97.48 & 97.54 \\
CIFAR100 & 97.5\% & 2 & 97.50 & 97.46 & 97.46 & 97.46 & 97.47 & 97.48 & 97.58 \\
\bottomrule
\end{tabular}
\caption{Actual sparsity (\%) on ResNet34. All methods achieve
comparable actual sparsity under the same target sparsity.}
\label{tab:app_resnet34_actual_sparsity_full}
\end{table*}

\section{Additional Dataset Checks}
\label{app:additional_datasets}

We evaluated 95\% sparsity on several transfer-style datasets to test
how broadly the recoverability transition pattern generalizes. These
experiments are not used as main evidence for RR dominance. Instead,
they show that the strongest allocation rule can be dataset dependent.
RR remains substantially above uniform layerwise pruning across all
transfer-style datasets and is often competitive with ERK, but global
magnitude pruning or raw activation diagnostics can be stronger on some
transfer-style datasets.

\begin{table*}[t]
\centering
\small
\begin{tabular}{lccccccc}
\toprule
Dataset & Seeds & Global & Uniform & Raw Shift & Repair Residual & RR & ERK \\
\midrule
Imagenette  & 3 & \textbf{75.84} & 37.27 & 72.19 & 72.19 & 71.23 & 69.55 \\
Imagewoof   & 1 & 55.13 & 29.60 & \textbf{55.28} & 55.18 & 52.79 & 52.25 \\
STL10       & 1 & \textbf{59.39} & 26.99 & 55.06 & 56.26 & 52.63 & 54.55 \\
Oxford Pets & 3 & \textbf{30.34} & 12.61 & 30.18 & 29.91 & 28.91 & 29.36 \\
\bottomrule
\end{tabular}
\caption{Additional 95\% sparsity checks on transfer-style datasets.
The strongest allocation rule varies across datasets. Imagenette and
Oxford Pets are averaged over three seeds; Imagewoof and STL10 are
single-seed. These results support the limitation that RR is a
transition-regime diagnostic rather than a universal allocation rule.}
\label{tab:app_transfer}
\end{table*}

On Imagenette, global magnitude pruning remains unexpectedly strong
after CR+BN, while RR is competitive with activation diagnostics and
improves over ERK on average. On Imagewoof and STL10, raw activation
diagnostics or global magnitude pruning are stronger than RR. On Oxford
Pets, global pruning, raw shift, repair residual, and ERK are within one
point of each other, while RR is slightly lower. These results suggest
that recoverability is not determined by sparsity alone; it may also
depend on dataset complexity, transfer gap, and feature distribution. We
leave a systematic study of these factors to future work.

\section{Calibration Sensitivity}
\label{app:cal_sensitivity}

RR uses calibration activations to estimate both raw pruning-induced
distortion and post-repair residual distortion. We therefore test
whether the allocation is sensitive to the number of calibration
examples. Table~\ref{tab:app_cal_sensitivity} reports CIFAR100 ResNet18
results at 95\% sparsity for calibration sizes from 16 to 128 images,
using seed 0. Across this range, RR remains stable and consistently
improves over ERK after CR+BN. Recovered accuracy varies by less than
0.4 points, and the RR minus ERK gap remains between \(+3.13\) and
\(+3.53\) points. The result indicates that the diagnostic does not
require a large calibration set to provide a useful allocation signal.

\begin{table}[t]
\centering
\small
\setlength{\tabcolsep}{5pt}
\begin{tabular}{ccccc}
\toprule
\(|\mathcal{C}|\) & RR & ERK & RR \( - \) ERK & RR sparsity \\
\midrule
16  & 12.51 & 9.10 & \(+3.41\) & 94.99 \\
32  & 12.76 & 9.23 & \(+3.53\) & 94.99 \\
64  & 12.42 & 9.29 & \(+3.13\) & 95.04 \\
128 & 12.52 & 9.31 & \(+3.21\) & 95.04 \\
\bottomrule
\end{tabular}
\caption{Calibration set size sensitivity. CIFAR100 ResNet18 at 95\%
sparsity, seed 0, accuracy after CR+BN. Recovered accuracy varies by
less than 0.4 points across calibration sizes from 16 to 128, and RR
remains 3.13 to 3.53 points above ERK throughout.}
\label{tab:app_cal_sensitivity}
\end{table}

\end{document}